\pdfoutput=1

\documentclass[11pt]{article}

\usepackage[preprint]{acl}

\usepackage{times}
\usepackage{latexsym}

\usepackage{float}
\usepackage{microtype}
\usepackage{hyperref}
\usepackage{url}
\usepackage{booktabs}
\usepackage{mathtools}
\usepackage{amsthm}
\usepackage{dsfont}
\usepackage{color}
\usepackage{multirow}
\usepackage{booktabs}
\usepackage{colortbl}
\usepackage{arydshln}
\usepackage{xcolor}
\usepackage{rotating}
\usepackage{subcaption}
\usepackage{longtable}

\usepackage{amsmath,amsfonts,bm}

\def\eqref#1{equation~\ref{#1}}

\def\1{\bm{1}}

\def\rt{{\textnormal{t}}}

\def\rx{{\textnormal{x}}}

\def\vphi{{\bm{\phi}}}
\def\vtheta{{\bm{\theta}}}
\def\va{{\bm{a}}}

\def\vd{{\bm{d}}}

\def\vq{{\bm{q}}}

\def\mQ{{\bm{Q}}}

\DeclareMathAlphabet{\mathsfit}{\encodingdefault}{\sfdefault}{m}{sl}
\SetMathAlphabet{\mathsfit}{bold}{\encodingdefault}{\sfdefault}{bx}{n}

\def\gD{{\mathcal{D}}}

\def\gH{{\mathcal{H}}}

\def\gP{{\mathcal{P}}}

\def\gV{{\mathcal{V}}}

\def\sP{{\mathbb{P}}}

\def\sR{{\mathbb{R}}}

\newcommand{\E}{\mathbb{E}}

\theoremstyle{definition}
\newtheorem{definition}{Definition}[section]

\usepackage[T1]{fontenc}

\usepackage[utf8]{inputenc}

\usepackage{microtype}

\usepackage{inconsolata}

\usepackage{graphicx}

\title{Do You Know What You Are Talking About? Characterizing Query-Knowledge Relevance For Reliable Retrieval Augmented Generation}

\author{
 \textbf{Zhuohang Li\textsuperscript{1}},
 \textbf{Jiaxin Zhang\textsuperscript{2}},
 \textbf{Chao Yan\textsuperscript{3}},
 \textbf{Kamalika Das\textsuperscript{2}},
\\
 \textbf{Sricharan Kumar\textsuperscript{2}},
 \textbf{Murat Kantarcioglu\textsuperscript{4}},
 \textbf{Bradley A. Malin\textsuperscript{1,3}}
\\
\\
 \textsuperscript{1}Vanderbilt University,
 \textsuperscript{2}Intuit AI Research,\\
 \textsuperscript{3}Vanderbilt University Medical Center,
 \textsuperscript{4}Virginia Tech
\\
\texttt{zhuohang.li@vanderbilt.edu},\\
\texttt{\{jiaxin\_zhang, kamalika\_das, sricharan\_kumar\}@intuit.com}\\
\texttt{\{chao.yan.1, b.malin\}@vumc.org}, \texttt{muratk@vt.edu}
}

\begin{document}
\maketitle
\begin{abstract}
Language models (LMs) are known to suffer from hallucinations and misinformation. Retrieval augmented generation (RAG) that retrieves verifiable information from an external knowledge corpus to complement the parametric knowledge in LMs provides a tangible solution to these problems. However, the generation quality of RAG is highly dependent on the relevance between a user's query and the retrieved documents. Inaccurate responses may be generated when the query is outside of the scope of knowledge represented in the external knowledge corpus or if the information in the corpus is out-of-date. In this work, we establish a statistical framework that assesses how well a query can be answered by an RAG system by capturing the relevance of knowledge. We introduce an online testing procedure that employs goodness-of-fit (GoF) tests to inspect the relevance of each user query to detect out-of-knowledge queries with low knowledge relevance. Additionally, we develop an offline testing framework that examines a collection of user queries, aiming to detect significant shifts in the query distribution which indicates the knowledge corpus is no longer sufficiently capable of supporting the interests of the users. We demonstrate the capabilities of these strategies through a systematic evaluation on eight question-answering (QA) datasets, the results of which indicate that the new testing framework is an efficient solution to enhance the reliability of existing RAG systems.
\end{abstract}

\section{Introduction}
\label{sec:intro}

Recent progress on large-scale pre-trained language models (LMs) \cite{brown2020language,anil2023palm} has demonstrated great potential in revolutionizing a wide array of applications across fields, ranging from natural language understanding and generation to complex problem-solving in scientific research.
Despite their remarkable abilities, generative LMs suffer from poor interpretability and transparency, as well as the intrinsic risk of hallucination and misinformation, which collectively prohibit them from being deployed in safety-critical domains such as healthcare \cite{wornow2023shaky,d2024large}.

Retrieval augmented generation (RAG) \cite{lewis2020retrieval} is a promising approach for enhancing language models (LMs) by incorporating verifiable, current information from external knowledge databases. Incorporating this external context to complement the inherent knowledge of LMs has demonstrated notable benefits in reducing occurrences of hallucination and misinformation, thereby improving the reliability of content produced \cite{shuster2021retrieval}. Still, numerous studies \cite{karpukhin2020dense,gao2022precise, tan2022tegtok,yan2024corrective} show that the effectiveness of RAG is dependent on the relevance between the query and retrieved documents. In cases where documents of weak relevance are provided, they can become distractions for the LM \cite{shi2023large}, leading to the generation of incorrect answers. At the present moment in time, there is no viable solution for safety-critical RAG systems to possess mechanisms for (1) evaluating the relevance of queries submitted from users to the knowledge corpus and flagging low-relevance queries in real-time that cannot be adequately addressed using the available knowledge or (2) identifying significant shifts in query distribution that are indicative of a potential misalignment between the knowledge corpus and user interests, which would suggest an outdated knowledge database that requires updating.

\begin{figure*}
    \centering
    \includegraphics[width=0.9\textwidth]{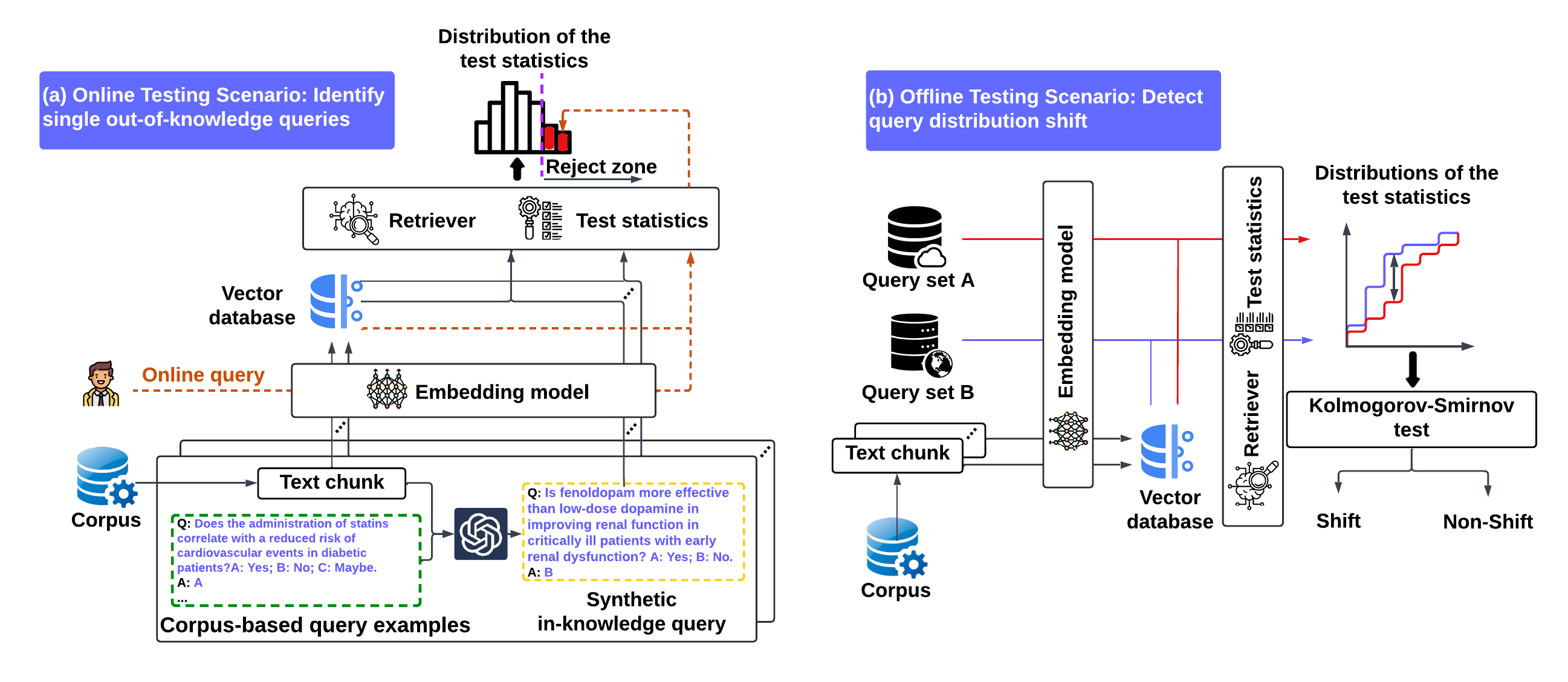}
    \vspace{-4mm}
    \caption{Overview of the hypothesis-testing framework for assessing query-knowledge relevance in RAG.}
    \vspace{-4mm}
    \label{fig:overview}
\end{figure*}

To address these deficiencies, in this paper we establish a statistical framework for accurate assessment of the query-knowledge relevance in retrieval-based LMs through hypothesis testing. As illustrated in Figure~\ref{fig:overview}, we introduce two testing procedures: an online testing procedure (left subfigure) that aims at identifying single out-of-knowledge queries and an offline testing procedure (right subfigure) for detecting query distribution shifts.
In the \textit{online} testing scenario, we cast evaluation of query-knowledge relevance as testing against the null hypothesis that the unknown query arises from the same distribution as the empirical in-knowledge queries (i.e., testing for goodness-of-fit (GoF)). We utilize the semantic similarity between the query and the retrieved most relevant documents captured by text embedding models \cite{izacardunsupervised} to derive test statistics and reject query samples that are unlikely to occur given the empirical distribution of in-knowledge queries (queries that can be answered with the knowledge corpus).
In scenarios when the in-knowledge query distribution is unknown, we generate synthetic in-knowledge queries, by prompting LMs with document chunks, to serve as a proxy to the true in-knowledge query distribution. In the \textit{offline} testing scenario, we employ a two-sample GoF test to determine whether the unknown set of queries matches the empirical distribution of historical in-knowledge queries, where a large $p$-value is suggestive of a significant query distribution shift.

To demonstrate the feasibility of these approaches, we report on a systematic evaluation with two biomedical corpus and eight QA datasets, including three
general domain QA datasets and five biomedical QA datasets, to investigate seven test statistics and six retrievers.
Our experiments highlight several notable findings. First, the testing-based methods can more reliably capture the relevance compared to LM-based relevance scores and outlier-detection-based baselines. Second, synthetic queries can provide a good approximate to the in-knowledge distribution with similar empirical performance for detecting out-of-knowledge queries.
Third, there is a misalignment between embedding models' ability to retrieve relevant documents and their ability to detect out-of-knowledge queries. And, fourth, query distribution shifts can be effectively detected through GoF testing with high accuracy using a relatively small sample size.

\vspace{-2mm}
\section{Background}
\label{sec:bg}
\vspace{-2mm}

\paragraph{Retrieval Augmented Generation (RAG).}
RAG systems~\cite{lewis2020retrieval} leverage external knowledge bases to assist language models (LMs) in responding to user queries. A RAG system is composed of a \textit{retriever} $\vphi$ and a \textit{generator} $\vtheta$. The retriever is connected to a \textit{corpus} of knowledge document chunks $\gD = \{\vd_i\}_{i=1}^{n}$, where $\vd_i \in \gV$ and $\gV$ denotes the space of natural texts. Given a user query $\vq \in \gV$, the retriever retrieves from $\gD$ the $k$-most relevant documents $\gD^r = \{\vd^r_i\}_{i=1}^k \subset \gD$. The generator is an LM that generates an answer $\va$ according to the query $\vq$ and the retrieved documents $\gD^r$. The overall framework can be described as
$\sP(\va|\vq, \gD)=\sP_\vphi(\gD^r|\vq)\sP_\vtheta(\va|\vq, \gD^r)$, where $\sP_\vphi(\gD^r|\vq)$ and $\sP_\vtheta(\va|\vq, \gD^r)$ denotes the retrieval and generation process respectively.

\vspace{-2mm}
\paragraph{Embedding Model.}
Most modern RAG systems utilize vector databases to construct retriever~\cite{gao2023retrieval}. In this setting, an embedding model $E_\vphi: \gV \rightarrow \sR^{m}$ is employed to encode the document chunks as a set of vector representations in a $m$-dimensional latent space that captures semantic similarity\footnote{Some frameworks use separate embedding models for the query and document.}. The retriever retrieves a document according to its measured similarity to the query, i.e., $\sP_\vphi(\vd|\vq)\propto \exp(S(E_\vphi(\vd), E_\vphi(\vq)))$, where $S: \sR^{m} \times \sR^{m} \rightarrow \sR$ is a similarity metric. Common choices of $S$ include cosine similarity and dot product.

\vspace{-2mm}
\paragraph{Goodness-of-Fit (GoF) Test.} A GoF test is a statistical procedure for comparing an observation $\rx$ to an expected distribution $\gP$. Formally, it decides between a null hypothesis $\gH_0: \rx \sim \gP$ indicative of fitness and an alternative hypothesis $\gH_1: \rx \not\sim \gP$. In a one-sided right-tailed test, the test statistic $\rt(\rx)$ is compared to the critical value $c$, and $\gH_0$ is rejected if $\rt(\rx)\geq c$. In practice, $c$ is calculated based on a pre-determined significant level $\alpha \coloneq \sP(\rt(\rx)\geq c | \gH_0)$ denoting the probability of falsely rejecting the null hypothesis when the null hypothesis is true. To help interpret the test result, a $p$-value is reported to indicate the probability of obtaining a test statistic that is equal to or more extreme than the actual observed value $t$ under the assumption of $\gH_0$, i.e., $p(t) \coloneq \sP(\rt(\rx)\geq t | \gH_0)$. By convention, $p(t) \leq \alpha$ is considered statistically significant to reject $\gH_0$.

\vspace{-2mm}
\section{A Statistical Characterization of Query-Knowledge Relevance}
\label{sec:method}
\vspace{-1mm}
\subsection{Problem Definition}
\vspace{-1mm}
In many RAG applications, it is important for the model developer or service provider to assess the effectiveness with which a query $\vq$ can be addressed using the existing corpus $\gD$.
This motivates the following definition that quantifies this level of effectiveness.

\begin{definition}[Query-Knowledge Relevance]\label{def:qkr}
    We define the relevance of a given query $\vq$ with respect to a corpus $\gD$ as
    \begin{equation}\label{eq:def}
    \setlength\abovedisplayskip{3pt}
    \setlength\belowdisplayskip{3pt}
        \begin{split}
        r(\vq|\gD) \coloneq & \sup_{\vtheta \in \bm{\Theta}} \big\{ \sup_{\gD^r \subseteq \gD} \E_{\va \sim \sP_\vtheta(\va|\vq, \gD^r)}[\mathds{1}_{\{\va = \va_{gt}\}}]\\
        & - \E_{\va \sim \sP_\vtheta(\va|\vq)}[\mathds{1}_{\{\va = \va_{gt}\}}] \big\},         
        \end{split}
    \end{equation}
    where $\va_{gt}$ is the ground truth answer and $\mathds{1}$ denotes the indicator function.
\end{definition}
According to Definition~\ref{def:qkr}, $r(\vq|\gD) \in [0, 1]$ and queries with higher query-knowledge relevance are more likely to be answered correctly with the knowledge presented in the corpus. We define in-knowledge query and out-of-knowledge query as follows:
\begin{definition}[In-Knowledge Query]
    $\vq$ is in-knowledge if $r(\vq|\gD)>0$.
\end{definition}
\begin{definition}[Out-of-Knowledge Query]
    $\vq$ is out-of-knowledge if $r(\vq|\gD)\leq0$.
\end{definition}

It should be noted that Equation~\ref{eq:def} cannot be computed in general due to the supremum over all document chunks and the expectation taken over the answer space. Instead, given a set of known in-knowledge queries (i.e., queries that are verified to be answerable with the corpus), we seek to capture the empirical relevance via a statistical test.

\vspace{-2mm}
\subsection{Online Testing Procedure For Identifying Single Out-of-Knowledge Query}
\vspace{-1mm}

In safety-critical domains with low fault tolerance, such as medicine and finance, out-of-knowledge queries should be detected in a timely manner to either be rejected or trigger human intervention to ensure output quality.
The following describes a testing procedure that decides whether a single query is out-of-knowledge in an online fashion.

\begin{definition}[GoF Test for Query Relevance]\label{def:test}
    Given a fixed in-knowledge query distribution $\gP_I$ and a new query $\vq$ sampled from unknown distribution $\gP$, the problem of deciding if $\vq$ is out-of-knowledge can be formalized as testing the simple null hypothesis
    \begin{equation*}
    \setlength\abovedisplayskip{3pt}
    \setlength\belowdisplayskip{3pt}
        \gH_0: \: \gP = \gP_I \quad \text{against} \quad
        \gH_1: \: \gP \neq \gP_I.
    \end{equation*}
\end{definition}

Let $F(t) = \sP(\rt(\vq)\leq t | \gH_0)$ denote the cumulative distribution function (CDF) of the test statistics under $\gH_0$. The the $p$-value of the test is $1 - F(t)$
and the critical value for a test of size $\alpha$ is $c(\alpha; \gP_I) \coloneq \inf \{t: F(t) > 1 - \alpha\}$. The GoF test then rejects $\gH_0$ if $1 - F(t) \leq \alpha$ or $t \geq c(\alpha; \gP_I)$, indicating that the sample is out-of-knowledge, with type I error bounded by $\alpha$.

In practice, the in-knowledge distribution is unknown, such that the $p$-value and critical value cannot be calculated analytically. However, they can be approximated through a sampling process.
Given a set of in-knowledge queries $\mQ_I$, we estimate the empirical cumulative distribution function (eCDF) of test statistics under $\gH_0$ using $\mQ_I$:
\begin{equation}
    \setlength\abovedisplayskip{3pt}
    \setlength\belowdisplayskip{3pt}
    \hat{F}(\rt(\vq); \mQ_I) = \frac{1 + \sum_{\vq_i \in \mQ_I} \mathds{1}_{\{\rt(\vq_i) \leq \rt(\vq)\}}}{1 + |\mQ_I|}.
\end{equation}
The $p$-value and critical value can thus be obtained as $p(\rt(\vq))= 1 - \hat{F}(\rt(\vq);\mQ_I)$ and $c(\alpha; \mQ_I) \coloneq \inf \{t: \hat{F}(t; \mQ_I) > 1 - \alpha\}$. According to Donsker's theorem, any desired precision of this estimation can be achieved by increasing the sample size.
A larger $p$-value indicates that the query is more relevant to the empirical distribution and thus more likely to be in knowledge.

\vspace{-2mm}
\paragraph{Test Statistics.}
The proposed hypothesis testing framework requires mapping a given test query to a numerical score as the test statistics. Ideally, the score should capture the degree to which the test query deviates from the distribution of queries encapsulated by the corpus, so that a higher score indicates that the test query is more likely to be out of knowledge.
As there is no clear guidance from prior literature on which test statistic provides the best performance in this scenario, we explore the following options:

\textit{(1) Maximum Similarity Score (MSS).} A simple baseline is to use the similarity score between the query and the most relevant document chunk from the corpus measured in the embedding space as the test statistic, i.e., $\rt(\vq)= - \max_{\vd \in \gD} S(E_\vphi(\vd), E_\vphi(\vq))$.

\textit{(2) $k$-th Nearest Neighbor (KNN).} An extension of the maximum similarity score is to estimate the similarity score between the query and its $k$-th nearest document embeddings \cite{ramaswamy2000efficient,sun2022out}, i.e., $\rt(\vq)= - s_k$ where $s_k$ is the $k$-th largest element of $\{S(E_\vphi(\vd), E_\vphi(\vq))| \vd \in \gD\}$.

\textit{(3) Average of $k$ Nearest Neighbors (AvgKNN).} AvgKNN computes the average similarity score of all $k$ nearest document embeddings \cite{angiulli2002fast} as the test statistic, i.e., $\rt(\vq)= - \frac{1}{k}\sum_{\vd \in \gD^r} S(E_\vphi(\vd), E_\vphi(\vq))$.

\textit{(4) Entropy.} The entropy of the retriever probability distribution $\sP_\vphi(\vd|\vq)$ captures the retrieval uncertainty where a higher value may be suggestive of potential out-of-knowledge queries \cite{ren2019likelihood}. For computational efficiency, we compute the entropy using only the $k$ nearest document entries, i.e., $\rt(\vq)= - \sum_{i=1}^k \sP(\vq_i) \log(\sP(\vq_i))$, where $\sP(\vq_i) = \frac{\exp(S(E_\vphi(\vd), E_\vphi(\vq_i))}{\sum_{\vq \in \gD^r}\exp(S(E_\vphi(\vd), E_\vphi(\vq))}$.

\textit{(5) Energy.} Prior work by~\cite{liu2020energy} suggests that the free energy function of a softmax-based neural classifier contains useful cues for distinguishing out-of-distribution samples. Here we extend this idea to embedding models trained with contrastive loss (similar to \cite{kim2022energy}) by computing the energy score of a query with respect to its $k$ nearest document entries as $\rt(\vq)=-\tau \log \sum_{i=1}^k g(\vq_i)$, where $g(\vq_i)=\exp(S(E_\vphi(\vd), E_\vphi(\vq_i)/\tau)$ and $\tau$ is a temperature parameter which we set to $1.0$ by default.

\vspace{-2mm}
\paragraph{Meta-analytic Testing.}
In addition to the aforementioned univariate tests, we conduct additional meta-analytic tests. This is accomplished by running $k$ independent univariate tests for each of the $k$ nearest neighbors and then performing a meta-analysis on the test results. We employ the Fisher \cite{fisher1970statistical} and Simes \cite{simes1986improved} methods for obtaining an aggregated test statistic for the global null hypothesis as suggested by \cite{haroush2022statistical}. Specifically, given a set of $p$-values $\{p_i(\vq)\}_{i=0}^k$ derived from performing $k$ independent tests using the $i$-th nearest neighbors, the \textit{(6) Fisher} method derives the test statistic as $-2\sum_{i=0}^k p_i(\vq)$ and the \textit{(7) Simes} method derives the test statistic as $\min_i \frac{k\cdot p_{(i)}}{i}$ where $p_{(i)}$ is the $i$-th $p$-value after sorting.

\vspace{-2mm}
\paragraph{Synthesizing In-Knowledge Queries.}
The above testing procedure requires obtaining a set of in-knowledge queries for estimating the eCDF. However, as noted, such a set of queries may not be readily available. Thus, we generate synthetic in-knowledge queries by prompting an LM to generate question-answer pairs based on each document chunk, i.e., by drawing samples from $\sP_\vtheta(\vq,\va|\gD)$. The answers generated along with the questions are intended to ensure the generated question can be answered based on the context of the corpus. We expect the obtained synthetic question set $\hat{\mQ_I}$ to serve as a proxy of the in-knowledge queries for deciding the critical region at development time and the derived threshold can be used at test time to detect out-of-knowledge queries. Note that with synthetic queries this test no longer has a bounded type I error rate
and the actual performance will depend on the divergence between $\hat{\mQ_I}$ and $\mQ_I$.

\vspace{-2mm}
\subsection{Offline Testing Procedure For Detecting Query Distribution Shift}
\vspace{-1mm}
We additionally consider an offline scenario where the service provider has the opportunity to review a collection of user queries gathered during the deployment stage of the knowledge base to identify any potential shifts in the query distribution.
This enables the service provider to determine if the RAG knowledge base needs to be updated in response to evolving user requirements.

We achieve this by extending the test defined in Definition~\ref{def:test} to test against multiple samples from the unknown distribution.
Specifically, given a set of in-knowledge queries $\mQ_I$ and a set of queries $\mQ_\gP$ from unknown distribution $\gP$, we employ the non-parametric two-sample Kolmogorov–Smirnov (KS) test that calculates the largest difference of the eCDFs as the test statistic, i.e.,
\begin{equation}
    \setlength\abovedisplayskip{3pt}
    \setlength\belowdisplayskip{3pt}
    t_{KS} = \sup_t | \hat{F}(t; \mQ_I) - \hat{F}(t; \mQ_\gP)|,
\end{equation}
where $t_{KS}$ follows the Kolmogorov distribution. We apply a two-tailed test which rejects the null hypothesis $\gH_0: \forall t, \hat{F}(t; \mQ_I) = \hat{F}(t; \mQ_\gP)$ at level $\alpha$ if $t_{KS} > \sqrt{-\ln(\frac{\alpha}{2})\cdot\frac{|\mQ_I|+|\mQ_\gP|}{2|\mQ_I|\cdot|\mQ_\gP|}}$.

\vspace{-2mm}
\section{Experiments}
\label{sec:exp}
\vspace{-2mm}
\subsection{Experimental Setup}
\vspace{-1mm}

\paragraph{Datasets and Corpora.}
We conduct experiments using queries from eight QA datasets, including three general domain QA datasets and five biomedical QA datasets from the MIRAGE benchmark \cite{xiong2024benchmarking}.
(1) \textit{TruthfulQA} \cite{lin2021truthfulqa} is a general domain QA dataset containing questions from 38 categories including law, finance, and politics. We select a subset with health-related questions excluded, resulting in a total number of $762$ questions.
(2) \textit{WikiQA} \cite{yang2015wikiqa} is a large set of $3,047$ general domain questions sampled from Bing query logs associated with Wikipedia pages.
(3) \textit{CommonsenseQA} \cite{talmor2018commonsenseqa} is a general domain QA dataset containing $12,247$ questions for testing common sense knowledge.
(4) \textit{MedQA-US} \cite{jin2021disease} is a medical examination QA dataset that includes $1,273$ multi-choice questions the US Medical Licensing Examination.
(5) \textit{MMLU-Med} \cite{hendrycks2021measuring} is a medical examination QA dataset containing $1,089$ questions selected from $6$ biomedicine related tasks from MMLU.
(6) \textit{MedMCQA} \cite{pal2022medmcqa} is a medical examination QA dataset with $4,183$ questions from Indian medical entrance exams.
(7) \textit{PubMedQA} \cite{jin2019pubmedqa} is a biomedical research QA dataset with $500$ questions that can be answered with yes/no/maybe indicative of the veracity of the statement based on scientific literature.
(8) \textit{BioASQ-Y/N} \cite{krithara2023bioasq} is a biomedical research QA dataset containing $618$ biomedical semantic questions from Task B of the BioASQ benchmark that can be answered with yes/no.

We consider two corpora from the biomedical domain as the knowledge base for the RAG system. The first is
\textit{Textbooks} \cite{jin2021disease}, which contains a collection of $18$ English medical textbooks. The second is \textit{PubMed} \cite{xiong2024benchmarking},
which contains abstracts from the biomedical literature.
MedQA-US serves as the ground truth in-knowledge queries for the Textbooks corpus and PubMedQA serves as the ground truth in-knowledge queries for the PubMed corpus, respectively, as both datasets are generated based on the corresponding corpus.
Unless otherwise indicated, we use Contriever \cite{izacardunsupervised} as the default embedding model for retriever in our experiments.

\vspace{-2mm}
\paragraph{Evaluation Metrics.}
The detection algorithms are evaluated on a balanced dataset (i.e., an equal number) of in-knowledge (IK) and out-of-knowledge (OoK) query samples.
We report on two threshold-independent metrics, namely, the area under the receiver operating characteristic curve (\textit{AUROC}) and the area under the precision-recall curve (\textit{AUPRC}). Additionally, we report the true positive rate (\textit{TPR}), i.e., the number of true out-of-knowledge samples over the total number of predicted out-of-knowledge samples, evaluated at a threshold that ensures the false positive rate on the in-knowledge queries reaches $5\%$. For testing-based methods, this implies a significance level of $5\%$.
We further report the detection error rate (\textit{DER}) at that threshold, i.e., the number of falsely classified queries over the total number of queries.

\begin{table*}
\centering
\caption{AUROC results of different test statistics.}
\label{tab:res_auroc}
\vspace{-2mm}
\begin{subtable}[h]{0.49\textwidth}
   \caption{\textit{Textbooks} Corpus}
   \label{subtab:res_auroc_textbooks}
   \vspace{-1mm}
    \setlength{\extrarowheight}{0pt}
    \addtolength{\extrarowheight}{\aboverulesep}
    \addtolength{\extrarowheight}{\belowrulesep}
    \setlength{\aboverulesep}{0pt}
    \setlength{\belowrulesep}{0pt}
    \resizebox{\linewidth}{!}{
    \begin{tabular}{c|ccccccc} 
    \toprule
    \multirow{2}{*}{\textbf{Dataset}}                          & \multicolumn{7}{c}{\textbf{Test Statistics}}                            \\ 
    \cline{2-8}
                                                       & MSS    & KNN    & AvgKNN & Entropy & Energy & Fisher & Simes   \\ 
    \midrule
    MMLU-US                                           & 0.8544 & 0.8563 & 0.8594 & 0.7884  & \textbf{0.8595} & 0.8588 & 0.8567  \\
    MedMCQA                                           & 0.9456 & 0.9430 & 0.9488 & 0.8737  & \textbf{0.9490} & 0.9481 & 0.9481  \\
    {\cellcolor[rgb]{0.94,0.94,0.94}}PubMedQA      & \textbf{0.9616} & 0.9451 & 0.9563 & 0.9251  & 0.9566 & 0.9554 & 0.9576  \\
    {\cellcolor[rgb]{0.94,0.94,0.94}}BioASQ-Y/N    & 0.9680 & 0.9700 & 0.9739 & 0.8964  & 0.9740 & 0.9734 & \textbf{0.9748}  \\
    {\cellcolor[rgb]{0.86,0.86,0.86}}TruthfulQA    & 0.9998 & \textbf{0.9999} & \textbf{0.9999} & 0.9647  & \textbf{0.9999} & 0.9994 & 0.9987  \\
    {\cellcolor[rgb]{0.86,0.86,0.86}}WikiQA        & 0.9981 & \textbf{0.9986} & \textbf{0.9986} & 0.9636  & \textbf{0.9986} & 0.9981 & 0.9975  \\
    {\cellcolor[rgb]{0.86,0.86,0.86}}CommonsenseQA & \textbf{0.9999} & \textbf{0.9999} & \textbf{0.9999} & 0.9710  & \textbf{0.9999} & 0.9994 & 0.9988  \\
    \bottomrule
    \end{tabular}
    }
\end{subtable}
\hfill
\begin{subtable}[h]{0.49\textwidth}
   \caption{\textit{PubMed} Corpus}
   \label{subtab:res_auroc_textbooks}
   \vspace{-1mm}
    \setlength{\extrarowheight}{0pt}
    \addtolength{\extrarowheight}{\aboverulesep}
    \addtolength{\extrarowheight}{\belowrulesep}
    \setlength{\aboverulesep}{0pt}
    \setlength{\belowrulesep}{0pt}
    \resizebox{\linewidth}{!}{
    \begin{tabular}{c|ccccccc}
    \toprule
    \multirow{2}{*}{\textbf{Dataset}}                          & \multicolumn{7}{c}{\textbf{Test Statistics}}                            \\ 
    \cline{2-8}
                                          & MSS    & KNN    & AvgKNN & Entropy & Energy & Fisher & Simes   \\ 
    \midrule
     BioASQ-Y/N                                     & \textbf{0.6567} & 0.5924 & 0.5814 & 0.5818  & 0.5818 & 0.5792 & 0.5876  \\
     {\cellcolor[rgb]{0.94,0.94,0.94}}MMLU-US       & \textbf{0.7440} & 0.6818 & 0.6857 & 0.6696  & 0.6867 & 0.6870 & 0.6883  \\
     {\cellcolor[rgb]{0.94,0.94,0.94}}MedMCQA       & \textbf{0.8292} & 0.7343 & 0.7424 & 0.7251  & 0.7439 & 0.7429 & 0.7486  \\
     {\cellcolor[rgb]{0.86,0.86,0.86}}TruthfulQA    & \textbf{0.9980} & 0.9973 & 0.9977 & 0.6956  & 0.9977 & 0.9961 & 0.9922  \\
     {\cellcolor[rgb]{0.86,0.86,0.86}}WikiQA        & \textbf{0.9937} & 0.9906 & 0.9910 & 0.7333  & 0.9911 & 0.9894 & 0.9859  \\
     {\cellcolor[rgb]{0.86,0.86,0.86}}CommonsenseQA & \textbf{0.9994} & 0.9985 & 0.9989 & 0.7507  & 0.9989 & 0.9969 & 0.9928  \\
    \bottomrule
    \end{tabular}
    }
\end{subtable}
\vspace{-2mm}
\end{table*}

\begin{table*}
\centering
\setlength{\extrarowheight}{0pt}
\addtolength{\extrarowheight}{\aboverulesep}
\addtolength{\extrarowheight}{\belowrulesep}
\setlength{\aboverulesep}{0pt}
\setlength{\belowrulesep}{0pt}
\caption{Comparison with outlier detection-based baselines on the Textbooks corpus.}
\label{tab:baseline}
\vspace{-1mm}
\resizebox{0.94\linewidth}{!}{
\begin{tabular}{@{\extracolsep{4pt}}c|cccccccccc:cc@{}} 
\toprule
\multirow{2}{*}{\textbf{Dataset }} & \multicolumn{2}{c}{\textbf{Maha }} & \multicolumn{2}{c}{\textbf{SVM }} & \multicolumn{2}{c}{\textbf{LOF }} & \multicolumn{2}{c}{\textbf{KDE }} & \multicolumn{2}{c:}{\textbf{COPOD }} & \multicolumn{2}{c}{\textbf{GoF {\footnotesize(Energy)}}}  \\ 
\cline{2-3} \cline{4-5} \cline{6-7} \cline{8-9} \cline{10-11} \cline{12-13}
                                   & AUROC  & AUPRC                     & AUROC  & AUPRC                    & AUROC  & AUPRC                    & AUROC  & AUPRC                    & AUROC  & AUPRC                       & AUROC           & AUPRC            \\ 
\midrule
MMLU-US                            & 0.7700 & 0.7773                    & 0.7843 & 0.7774                   & 0.7600 & 0.7674                   & 0.7491 & 0.7444                   & 0.6265 & 0.6537                      & \textbf{0.8595} & \textbf{0.8741}  \\
MedMCQA                            & 0.7681 & 0.7422                    & 0.8192 & 0.7756                   & 0.7590 & 0.7279                   & 0.7627 & 0.7250                   & 0.6209 & 0.5902                      & \textbf{0.9490} & \textbf{0.9436}  \\
PubMedQA                           & 0.9274 & 0.8957                    & 0.9402 & 0.8975                   & 0.9165 & 0.8554                   & 0.9145 & 0.8536                   & 0.8222 & 0.7291                      & \textbf{0.9566} & \textbf{0.9458}  \\
BioASQ-Y/N                         & 0.9076 & 0.8703                    & 0.9376 & 0.8944                   & 0.9151 & 0.8586                   & 0.9110 & 0.8464                   & 0.8143 & 0.7172                      & \textbf{0.9740} & \textbf{0.9660}  \\
TruthfulQA                         & 0.8001 & 0.6941                    & 0.8290 & 0.7046                   & 0.7424 & 0.6191                   & 0.7684 & 0.6362                   & 0.7425 & 0.6191                      & \textbf{0.9999} & \textbf{0.9999}  \\
WikiQA                             & 0.7244 & 0.6361                    & 0.7704 & 0.6582                   & 0.6704 & 0.5735                   & 0.6992 & 0.5901                   & 0.6541 & 0.5605                      & \textbf{0.9986} & \textbf{0.9986}  \\
CommonsenseQA                      & 0.7179 & 0.6044                    & 0.7601 & 0.6284                   & 0.6402 & 0.5330                   & 0.6893 & 0.5665                   & 0.6629 & 0.5495                      & \textbf{0.9999} & \textbf{0.9999}  \\
\bottomrule
\end{tabular}
}
\vspace{-1mm}
\end{table*}

\vspace{-2mm}
\subsection{Online Testing Results}
\vspace{-1mm}

\paragraph{Comparison of Test Statistics.}
We compare test statistics
by randomly drawing $300$ samples from the IK dataset and $300$ samples from the OoK dataset to construct a balanced testing set and measuring the AUROC and AUPRC in detecting OoK queries.
We set $k=32$ and report the average result over $10$ independent runs for stability.
Table~\ref{tab:res_auroc} presents the AUROC, while Table~\ref{tab:res_auprc} in Appendix reports on the AUPRC results due to space limits.
For each corpus, the OoK datasets are labeled with bright colors indicating near OoK and dark colors indicating far OoK.
We observe that on both corpora, all test statistics achieve high performance in distinguishing far OoK queries, with many achieving over $0.99$ AUROC on both corpora. We additionally find that the performance of test statistics varies from each corpus. For instance, Energy achieves the overall best AUROC on the Textbooks corpus while MSS performs the best among all test statistics on the PubMed corpus.
This suggests the optimal test statistic should be decided for each application domain.

\vspace{-2mm}
\paragraph{Comparison with Outlier Detection-based Baselines.} We compare the GoF test using energy scores with five common outlier detection algorithms,
including \textit{Mahalanobis distance (Maha)} \cite{hardin2004outlier,lee2018simple}, \textit{One-class SVM (SVM)} \cite{scholkopf2001estimating}, \textit{Local Outlier Factor (LOF)} \cite{breunig2000lof}, \textit{Kernel Density Estimation (KDE)} \cite{latecki2007outlier}, and \textit{Copula-Based Outlier Detection (COPOD)} \cite{li2020copod}. Based on the results presented in Table~\ref{tab:baseline}, it can be seen that the GoF test consistently achieves the best result across all datasets. The weaker performance of the baseline algorithms is likely due to the limited sample size and the intrinsic difficulty in density modeling for high-dimensional data. Notably, this result demonstrates that GoF testing is more sample-efficient for detecting out-of-knowledge queries compared with conventional outlier detection algorithms.

\vspace{-2mm}
\paragraph{Comparison with LM-based Relevance Score.}
In Table~\ref{tab:lm_score}, we compare the GoF test with LM-based relevance scores. Specifically, we ask the LM to generate a numerical relevance score for each query $\vq$ and its retrieved relevant documents $\gD^r$ using the prompt in Table~\ref{tab:lm_score_prompt}. It can be seen that the LM-based score failed to capture the relevance between the query and the knowledge corpus, resulting in poor performance in detecting OoK queries. This is potentially because of LM's lack of ability to produce accurate numerical scores ~\cite{spithourakis2018numeracy,liu2023g} and its tendency for hallucination when processing complex concepts in long texts \cite{ji2023survey}. We provide examples of hallucinated responses in Appendix Table~\ref{tab:hallucinated_LM_score}.

\begin{table}
\centering
\caption{Comparison with LM-based relevance score on the Textbooks corpus.}
\label{tab:lm_score}
\vspace{-2mm}
\setlength{\extrarowheight}{0pt}
\addtolength{\extrarowheight}{\aboverulesep}
\addtolength{\extrarowheight}{\belowrulesep}
\setlength{\aboverulesep}{0pt}
\setlength{\belowrulesep}{0pt}
\resizebox{\linewidth}{!}{
\begin{tabular}{@{\extracolsep{4pt}}c|cccc:cc@{}} 
\toprule
\multirow{2}{*}{\textbf{Dataset}} & \multicolumn{2}{c}{\textbf{GPT-3.5}} & \multicolumn{2}{c:}{\textbf{GPT-4}} & \multicolumn{2}{c}{\textbf{GoF {\footnotesize(Energy)}}}  \\ 
\cline{2-3} \cline{4-5} \cline{6-7}
                  & AUROC  & AUPRC              & AUROC  & AUPRC            & AUROC  & AUPRC           \\ 
\hline
PubMedQA          & 0.1623 & 0.4161             & 0.2088 & 0.4078           & \textbf{0.9566} & \textbf{0.9458}          \\
CommonsenseQA     & 0.4975 & 0.7218             & 0.5001 & 0.6788           & \textbf{0.9999} & \textbf{0.9999}          \\
\bottomrule
\end{tabular}
}
\vspace{-2mm}
\end{table}

\begin{figure*}
    \centering
    \includegraphics[width=0.88\linewidth]{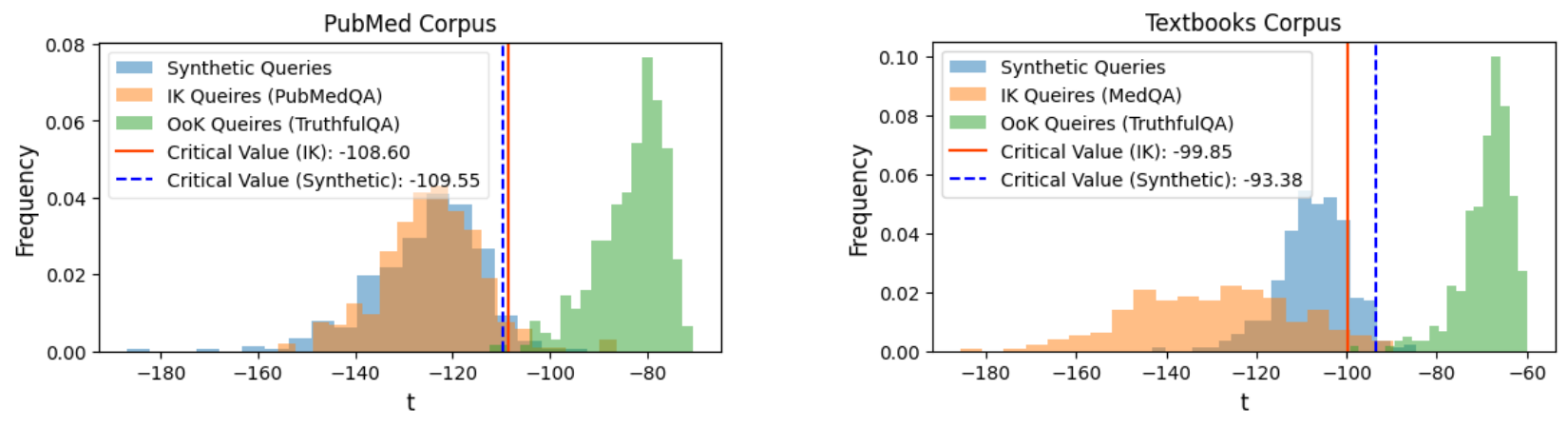}
    \vspace{-4mm}
    \caption{Illustration of critical values estimated using true in-knowledge and synthetic queries and histograms on the Textbooks corpus with energy score as test statistic.}
    \label{fig:hist}
    \vspace{-2mm}
\end{figure*}

\vspace{-2mm}
\paragraph{Synthetic Queries.} The results for threshold-independent metrics assumed that the true IK query distribution is known. However, this is not the case in practice and, thus, we relax this assumption by comparing the TPR and DER results produced with critical values estimated using true IK queries with results produced with synthetic queries.
We use \texttt{gpt-3.5-turbo} as the LM for generating synthetic queries and include the prompt and samples of generated synthetic queries in the Appendix.
Table~\ref{tab:synthetic} presents the results of using synthetic queries on the TruthfulQA datasets and Figure~\ref{fig:hist} depicts the histograms on both corpora.

It can be seen that the synthetic query distribution on the PubMed corpus closely matches the true IK query distribution, resulting in similar estimations of the critical region. Consequently, the differences between TPRs and DERs produced with IK queries and synthetic queries are negligible.
On the Textbooks corpus, however, we find that the synthetic query distribution deviates slightly from the true IK query distribution, which is likely due to the complexity of questions in MedQA. This results in a more conservative estimation of the critical values (more towards the right). Nevertheless, the performance of synthetic queries is still on par with true IK queries in terms of distinguishing far OoK queries.
Along the test statistics dimension, all test statistics are able to effectively distinguish OoK queries using synthetic queries except for Entropy which failed on the PubMed corpus and produces suboptimal results on the Textbooks corpus.
We include an extended version of Table~\ref{tab:synthetic} and Figure~\ref{fig:hist} in the Appendix.

\begin{table*}
\centering
\caption{Comparison of true positive rate (TPR) and detection error rate (DER) with critical values estimated using true IK and synthetic queries on TruthfulQA.}
\label{tab:synthetic}
\vspace{-1mm}
\setlength{\extrarowheight}{0pt}
\addtolength{\extrarowheight}{\aboverulesep}
\addtolength{\extrarowheight}{\belowrulesep}
\setlength{\aboverulesep}{0pt}
\setlength{\belowrulesep}{0pt}
\resizebox{0.78\linewidth}{!}{
\begin{tabular}{c|c|c|ccccccc} 
\toprule
\multirow{2}{*}{\textbf{Corpus}}    & \multirow{2}{*}{\textbf{Data Source}}    & \multirow{2}{*}{\begin{tabular}[c]{@{}l@{}}\textbf{Metric}\\$\alpha=5\%$\end{tabular}}                          & \multicolumn{7}{c}{\textbf{Test Statistics}}                            \\ 
\cline{4-10}
                        &                                       &           & MSS    & KNN    & AvgKNN & Entropy & Energy & Fisher & Simes   \\ 
\midrule
\multirow{4}{*}{PubMed} & \multirow{2}{*}{In-knowledge Queries} & TPR   & 0.9960 & 0.9960 & 0.9966 & 0.0930  & 0.9973 & 0.9976 & 0.9956 \\
                        &                                       & DER       & 0.0251 & 0.0213 & 0.0244 & 0.4786  & 0.0253 & 0.0321 & 0.0249  \\
\cline{2-10}
                        & \multirow{2}{*}{Synthetic Queries}    & TPR       & 0.9963 & 0.9973 & 0.9973 & 0.1410  & 0.9976 & 0.9976 & 0.9960  \\
                        &                                       & DER       & 0.0273 & 0.0268 & 0.0278 & 0.4623  & 0.0286 & 0.0441 & 0.0265  \\
\midrule
\multirow{4}{*}{Textbooks} & \multirow{2}{*}{In-knowledge Queries} & TPR   & 0.9993 & 1.0    & 1.0    & 0.8160  & 1.0    & 1.0    & 0.9996 \\
                        &                                       & DER       & 0.0153 & 0.0088 & 0.0101 & 0.1186  & 0.0116 & 0.0471 & 0.0241  \\
\cline{2-10}
                        & \multirow{2}{*}{Synthetic Queries}    & TPR       & 0.9990 & 0.9990 & 0.9990 & 0.6446  & 0.9990 & 0.9990 & 0.9990  \\
                        &                                       & DER       & 0.0081 & 0.0025 & 0.0035 & 0.1861  & 0.0039 & 0.0181 & 0.0069  \\
\bottomrule
\end{tabular}
}
\vspace{-3mm}
\end{table*}

\begin{figure}[t]
    \centering
    \includegraphics[width=0.8\linewidth]{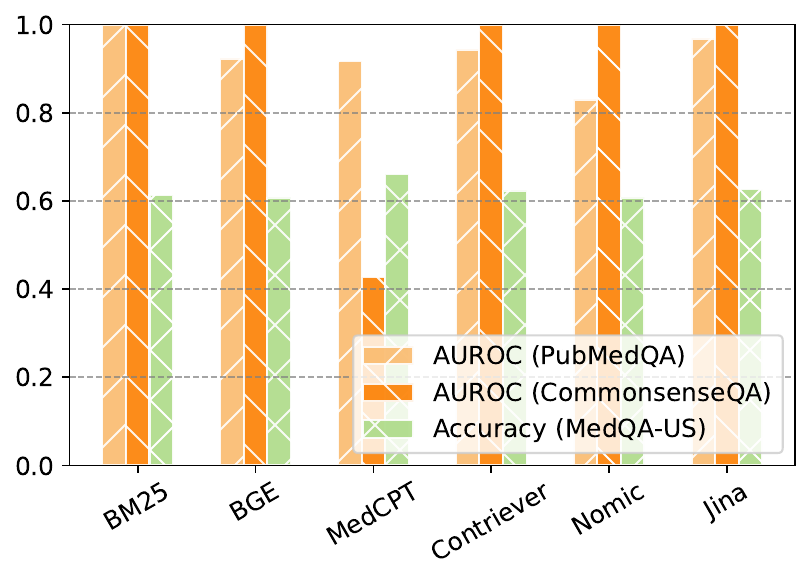}
    \vspace{-4mm}
    \caption{Comparison of six different embedding models.}
    \label{fig:retriever}
    \vspace{-4mm}
\end{figure}

\begin{figure}[t]
    \centering
    \includegraphics[width=0.8\linewidth]{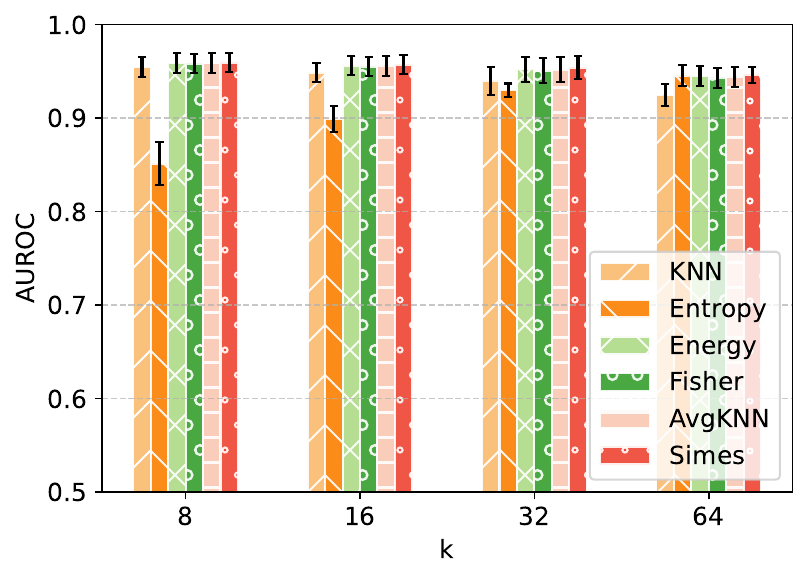}
    \vspace{-4mm}
    \caption{AUROC for PubMedQA as a function of $k$.}
    \label{fig:varying_k}
    \vspace{-4mm}
\end{figure}

\begin{figure*}[t]
    \centering
    \includegraphics[width=0.8\linewidth]{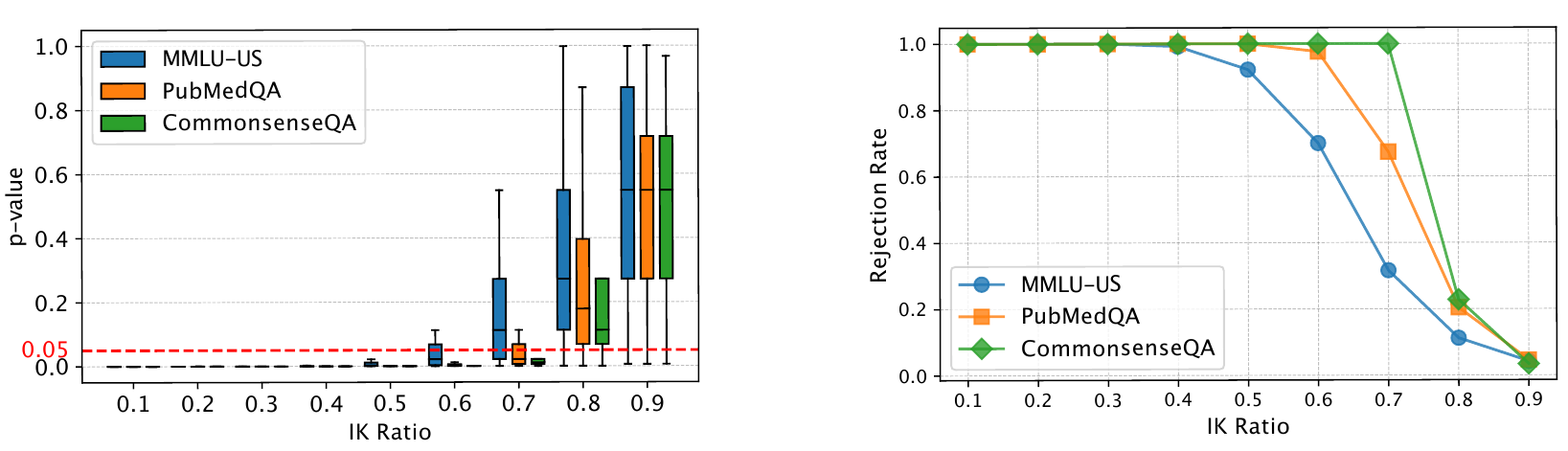}
    \vspace{-2mm}
    \caption{Offline testing results with the Textbooks corpus.}
    \label{fig:offline}
    \vspace{-3mm}
\end{figure*}

\vspace{-2mm}
\paragraph{Comparison of Embedding Models.}
We compare six embedding models including BM25~\cite{robertson2009probabilistic}, BGE~\cite{xiao2023c}, Contriever~\cite{izacardunsupervised}, MedCPT~\cite{jin2023medcpt}, Nomic~\cite{nussbaum2024nomic}, and Jina~\cite{gunther2023jina}, in terms of their ability to distinguish IK and OoK queries and their ability to retrieve relevant documents.
In Figure~\ref{fig:retriever}, we plot the AUROC of detecting OoK queries with KNN test statistics on the Textbooks dataset and the accuracy of the RAG system answering multi-choice questions from the IK dataset (MedQA-US), using \texttt{gpt-3.5-turbo} as the LM.
We have the following observations: (1) Different embedding models show disparate impact on detecting OoK queries; (2) The embedding model's performance of OoK query detection does not align with its performance of retrieving relevant documents. For example, MedCPT, a domain-specific embedding model pre-trained on biomedical data, shows the best QA accuracy on the MedQA-US dataset but has the lowest AUROC in detecting OoK queries. Additionally, it appears that pre-training on domain-specific data has a negative impact on the embedding model's ability to distinguish general queries, resulting in a lower AUROC on CommonsenseQA compared to PubMedQA, which is also evident from Figure~\ref{fig:retriever_hist} in Appendix.

\vspace{-2mm}
\paragraph{Sensitivity Analysis for $k$.}
In Figure~\ref{fig:varying_k}, we report on experiments with the Textbooks corpus across a range of $k$ values to study the impact of the number of retrieved samples on the performance of detecting OoK queries.
As $k$ increases from $8$ to $64$, it can be seen that the AUROC of Entropy improves from $0.8511$ to $0.9451$ and the AUROC of KNN decreases slightly from $0.9543$ to $0.9245$, while results on other test statistics remain stable across different $k$.

\vspace{-2mm}
\subsection{Offline Testing Results}
\vspace{-1mm}

We performed an evaluation of the offline testing procedure on the Textbooks corpus with three OoK datasets, i.e., MMLU-US, PubMedQA, and CommonsenseQA, using energy as the test statistic. To do so, we draw $50$ random samples from the IK dataset (MedQA-US) as $\mQ_I$ and another $50$ random samples from both the IK dataset and the OoK dataset to construct $\mQ_\gP$ with varying IK sample ratios. At each IK ratio, we repeat the experiments $500$ times and collect the $p$-values. We consider a confidence level of $0.05$, i.e., predict $\mQ_\gP$ to be OoK if the $p$-value is below $0.05$.
Figure~\ref{fig:offline} depicts the $p$-value distribution and the rejection rate (i.e., the number of occurrences predicting $\mQ_\gP$ as OoK over the total number of trials) at different IK ratios.
It can be seen that the general trend that $p$-values of the offline test decrease with the IK ratio, which is expected since it is easier to detect the distribution shift with more OoK samples in $\mQ_\gP$.
Similar to online testing, near OoK queries are harder to detect. For example, $\mQ_\gP$ containing only $30\%$ CommonsenseQA queries can be rejected at $100\%$ rate, whereas rejecting MMLU-US queries at $100\%$ rate requires at least $60\%$ OoK samples in $\mQ_\gP$.

\vspace{-2mm}
\section{Related Work}
\label{sec:related}
\vspace{-1mm}

\paragraph{Retrieval Augmented Generation.}
Retrieval augmented generation (RAG) \cite{lewis2020retrieval} provides a tangible solution to address hallucinations in knowledge-intensive tasks by retrieving verifiable information from external knowledge corpora \cite{shuster2021retrieval, yang2023alignment,wang2023self}.
However, the truthfulness of RAG responses highly depends on the relevance between the query and the corpus \cite{karpukhin2020dense,tan2022tegtok,yan2024corrective} and may suffer from increased risk of hallucination under distribution shifts \cite{kang2024c}.
While recent studies mostly focus on improving various stages of the retrieval and generation pipeline \cite{jiang2023active,yan2024corrective,asai2023self,yu2022generate}, in this work, we take a different perspective and improve the RAG systems' awareness of their knowledge boundary to indicate what they know and when are sufficiently certain that they should return knowledge to an end user \cite{ren2023investigating,kadavath2022language,yin2023large,ni2024llms}. 

\vspace{-2mm}
\paragraph{Out-of-distribution Detection.} Our work is related to the emerging research field of out-of-distribution (OoD) detection \cite{yang2021generalized}, which aims to detect test samples that are outside of the training data distribution. However, the large body of prior OoD detection research is in the vision domain with a focus on classification problems \cite{lee2018simple,liu2020energy} and only few recent work \cite{zhou2021contrastive,ren2022out} explore OoD detection for LMs.
Our work aims to address a different problem of identifying queries that are beyond the knowledge boundary of the corpus, where existing OoD detection algorithms are inapplicable as most embedding models are trained on general domain data.

\vspace{-2mm}
\section{Conclusion}
\label{sec:conclusion}
\vspace{-2mm}

Identifying out-of-knowledge queries is an important step in improving the reliability of RAG systems and reducing hallucination and misinformation. This work establishes a statistical framework for quantifying the relevance of query-knowledge relevance through goodness-of-fit hypothesis testing. We introduce two testing procedures with different goals of identifying low-relevance query samples and detecting query distribution shifts. We demonstrate the effectiveness of these approaches through extensive experiments on eight datasets from various domains. We hope our findings can provide insights for future research on reliable retrieval-based generation.

\pagebreak

\section*{Limintations}
A reliable RAG system should ensure both relevance in retrieval and faithfulness in generation. Our work focuses on the relevance between the user query and the existing knowledge database to abstain/reject queries with high risk (i.e., low relevance) and identify significant query distribution shift, which is an important prerequisite for reliable generation, but this may not provide the complete picture of the reliability of RAG in general.
In practice, our method can be deployed jointly with other approaches that aim to improve the reliability of LLMs in the generation phase (e.g., SelfCheckGPT~\cite{manakul2023selfcheckgpt}) to ensure end-to-end robustness. Our method also extends to other applications beyond reliable generation, for instance, improving RAG system efficiency by selecting the most relevant database when multiple data sources are available.

\section*{Ethics Statement}
The overarching goal of our work is to enhance the reliability of RAG systems, reduce the risk of misinformation, and improve system trustworthiness.
These improvements are crucial for the ethical and safe deployment of AI, as they help to mitigate risks associated with unreliable outputs and promote user trust. Data sets used in our experiments are sourced from the open domain and do not pose any ethical concerns.

\section*{Acknowledgments}

This research was supported in part by NSF awards DMS-2204795, OAC-2115094, CNS-2331424, ARL/Army Research Office awards W911NF-24-1-0202 and W911NF-24-2-0114, NIH awards 5RM1HG009034-08, 5U54HG012510-04, and K99LM014428, and the Intuit University Collaboration Program.

\bibliography{reference}

\begin{thebibliography}{61}
\providecommand{\natexlab}[1]{#1}

\bibitem[{Angiulli and Pizzuti(2002)}]{angiulli2002fast}
Fabrizio Angiulli and Clara Pizzuti. 2002.
\newblock Fast outlier detection in high dimensional spaces.
\newblock In \emph{European conference on principles of data mining and knowledge discovery}, pages 15--27. Springer.

\bibitem[{Anil et~al.(2023)Anil, Dai, Firat, Johnson, Lepikhin, Passos, Shakeri, Taropa, Bailey, Chen et~al.}]{anil2023palm}
Rohan Anil, Andrew~M Dai, Orhan Firat, Melvin Johnson, Dmitry Lepikhin, Alexandre Passos, Siamak Shakeri, Emanuel Taropa, Paige Bailey, Zhifeng Chen, et~al. 2023.
\newblock Palm 2 technical report.
\newblock \emph{arXiv preprint arXiv:2305.10403}.

\bibitem[{Asai et~al.(2023)Asai, Wu, Wang, Sil, and Hajishirzi}]{asai2023self}
Akari Asai, Zeqiu Wu, Yizhong Wang, Avirup Sil, and Hannaneh Hajishirzi. 2023.
\newblock Self-rag: Learning to retrieve, generate, and critique through self-reflection.
\newblock \emph{arXiv preprint arXiv:2310.11511}.

\bibitem[{Breunig et~al.(2000)Breunig, Kriegel, Ng, and Sander}]{breunig2000lof}
Markus~M Breunig, Hans-Peter Kriegel, Raymond~T Ng, and J{\"o}rg Sander. 2000.
\newblock Lof: identifying density-based local outliers.
\newblock In \emph{Proceedings of the 2000 ACM SIGMOD international conference on Management of data}, pages 93--104.

\bibitem[{Brown et~al.(2020)Brown, Mann, Ryder, Subbiah, Kaplan, Dhariwal, Neelakantan, Shyam, Sastry, Askell et~al.}]{brown2020language}
Tom Brown, Benjamin Mann, Nick Ryder, Melanie Subbiah, Jared~D Kaplan, Prafulla Dhariwal, Arvind Neelakantan, Pranav Shyam, Girish Sastry, Amanda Askell, et~al. 2020.
\newblock Language models are few-shot learners.
\newblock \emph{Advances in neural information processing systems}, 33:1877--1901.

\bibitem[{D’Antonoli et~al.(2024)D’Antonoli, Stanzione, Bluethgen, Vernuccio, Ugga, Klontzas, Cuocolo, Cannella, and Ko{\c{c}}ak}]{d2024large}
Tugba~Akinci D’Antonoli, Arnaldo Stanzione, Christian Bluethgen, Federica Vernuccio, Lorenzo Ugga, Michail~E Klontzas, Renato Cuocolo, Roberto Cannella, and Burak Ko{\c{c}}ak. 2024.
\newblock Large language models in radiology: fundamentals, applications, ethical considerations, risks, and future directions.
\newblock \emph{Diagnostic and Interventional Radiology}, 30(2):80.

\bibitem[{Fisher(1970)}]{fisher1970statistical}
Ronald~Aylmer Fisher. 1970.
\newblock Statistical methods for research workers.
\newblock In \emph{Breakthroughs in statistics: Methodology and distribution}, pages 66--70. Springer.

\bibitem[{Gao et~al.(2022)Gao, Ma, Lin, and Callan}]{gao2022precise}
Luyu Gao, Xueguang Ma, Jimmy Lin, and Jamie Callan. 2022.
\newblock Precise zero-shot dense retrieval without relevance labels.
\newblock \emph{arXiv preprint arXiv:2212.10496}.

\bibitem[{Gao et~al.(2023)Gao, Xiong, Gao, Jia, Pan, Bi, Dai, Sun, and Wang}]{gao2023retrieval}
Yunfan Gao, Yun Xiong, Xinyu Gao, Kangxiang Jia, Jinliu Pan, Yuxi Bi, Yi~Dai, Jiawei Sun, and Haofen Wang. 2023.
\newblock Retrieval-augmented generation for large language models: A survey.
\newblock \emph{arXiv preprint arXiv:2312.10997}.

\bibitem[{G{\"u}nther et~al.(2023)G{\"u}nther, Milliken, Geuter, Mastrapas, Wang, and Xiao}]{gunther2023jina}
Michael G{\"u}nther, Louis Milliken, Jonathan Geuter, Georgios Mastrapas, Bo~Wang, and Han Xiao. 2023.
\newblock Jina embeddings: A novel set of high-performance sentence embedding models.
\newblock \emph{arXiv preprint arXiv:2307.11224}.

\bibitem[{Hardin and Rocke(2004)}]{hardin2004outlier}
Johanna Hardin and David~M Rocke. 2004.
\newblock Outlier detection in the multiple cluster setting using the minimum covariance determinant estimator.
\newblock \emph{Computational Statistics \& Data Analysis}, 44(4):625--638.

\bibitem[{Haroush et~al.(2022)Haroush, Frostig, Heller, and Soudry}]{haroush2022statistical}
Matan Haroush, Tzviel Frostig, Ruth Heller, and Daniel Soudry. 2022.
\newblock A statistical framework for efficient out of distribution detection in deep neural networks.
\newblock In \emph{International Conference on Learning Representations}.

\bibitem[{Hendrycks et~al.(2021)Hendrycks, Burns, Basart, Zou, Mazeika, Song, and Steinhardt}]{hendrycks2021measuring}
Dan Hendrycks, Collin Burns, Steven Basart, Andy Zou, Mantas Mazeika, Dawn Song, and Jacob Steinhardt. 2021.
\newblock Measuring massive multitask language understanding.
\newblock In \emph{International Conference on Learning Representations}.

\bibitem[{Izacard et~al.(2022)Izacard, Caron, Hosseini, Riedel, Bojanowski, Joulin, and Grave}]{izacardunsupervised}
Gautier Izacard, Mathilde Caron, Lucas Hosseini, Sebastian Riedel, Piotr Bojanowski, Armand Joulin, and Edouard Grave. 2022.
\newblock Unsupervised dense information retrieval with contrastive learning.
\newblock \emph{Transactions on Machine Learning Research}.

\bibitem[{Ji et~al.(2023)Ji, Lee, Frieske, Yu, Su, Xu, Ishii, Bang, Madotto, and Fung}]{ji2023survey}
Ziwei Ji, Nayeon Lee, Rita Frieske, Tiezheng Yu, Dan Su, Yan Xu, Etsuko Ishii, Ye~Jin Bang, Andrea Madotto, and Pascale Fung. 2023.
\newblock Survey of hallucination in natural language generation.
\newblock \emph{ACM Computing Surveys}, 55(12):1--38.

\bibitem[{Jiang et~al.(2023)Jiang, Xu, Gao, Sun, Liu, Dwivedi-Yu, Yang, Callan, and Neubig}]{jiang2023active}
Zhengbao Jiang, Frank~F Xu, Luyu Gao, Zhiqing Sun, Qian Liu, Jane Dwivedi-Yu, Yiming Yang, Jamie Callan, and Graham Neubig. 2023.
\newblock Active retrieval augmented generation.
\newblock \emph{arXiv preprint arXiv:2305.06983}.

\bibitem[{Jin et~al.(2021)Jin, Pan, Oufattole, Weng, Fang, and Szolovits}]{jin2021disease}
Di~Jin, Eileen Pan, Nassim Oufattole, Wei-Hung Weng, Hanyi Fang, and Peter Szolovits. 2021.
\newblock What disease does this patient have? a large-scale open domain question answering dataset from medical exams.
\newblock \emph{Applied Sciences}, 11(14):6421.

\bibitem[{Jin et~al.(2019)Jin, Dhingra, Liu, Cohen, and Lu}]{jin2019pubmedqa}
Qiao Jin, Bhuwan Dhingra, Zhengping Liu, William~W Cohen, and Xinghua Lu. 2019.
\newblock Pubmedqa: A dataset for biomedical research question answering.
\newblock \emph{arXiv preprint arXiv:1909.06146}.

\bibitem[{Jin et~al.(2023)Jin, Kim, Chen, Comeau, Yeganova, Wilbur, and Lu}]{jin2023medcpt}
Qiao Jin, Won Kim, Qingyu Chen, Donald~C Comeau, Lana Yeganova, W~John Wilbur, and Zhiyong Lu. 2023.
\newblock Medcpt: Contrastive pre-trained transformers with large-scale pubmed search logs for zero-shot biomedical information retrieval.
\newblock \emph{Bioinformatics}, 39(11):btad651.

\bibitem[{Johnson et~al.(2019)Johnson, Douze, and J{\'e}gou}]{johnson2019billion}
Jeff Johnson, Matthijs Douze, and Herv{\'e} J{\'e}gou. 2019.
\newblock Billion-scale similarity search with gpus.
\newblock \emph{IEEE Transactions on Big Data}, 7(3):535--547.

\bibitem[{Kadavath et~al.(2022)Kadavath, Conerly, Askell, Henighan, Drain, Perez, Schiefer, Hatfield-Dodds, DasSarma, Tran-Johnson et~al.}]{kadavath2022language}
Saurav Kadavath, Tom Conerly, Amanda Askell, Tom Henighan, Dawn Drain, Ethan Perez, Nicholas Schiefer, Zac Hatfield-Dodds, Nova DasSarma, Eli Tran-Johnson, et~al. 2022.
\newblock Language models (mostly) know what they know.
\newblock \emph{arXiv preprint arXiv:2207.05221}.

\bibitem[{Kang et~al.(2024)Kang, G{\"u}rel, Yu, Song, and Li}]{kang2024c}
Mintong Kang, Nezihe~Merve G{\"u}rel, Ning Yu, Dawn Song, and Bo~Li. 2024.
\newblock C-rag: Certified generation risks for retrieval-augmented language models.
\newblock \emph{arXiv preprint arXiv:2402.03181}.

\bibitem[{Karpukhin et~al.(2020)Karpukhin, Oguz, Min, Lewis, Wu, Edunov, Chen, and Yih}]{karpukhin2020dense}
Vladimir Karpukhin, Barlas Oguz, Sewon Min, Patrick Lewis, Ledell Wu, Sergey Edunov, Danqi Chen, and Wen-tau Yih. 2020.
\newblock Dense passage retrieval for open-domain question answering.
\newblock In \emph{Proceedings of the 2020 Conference on Empirical Methods in Natural Language Processing (EMNLP)}. Association for Computational Linguistics.

\bibitem[{Kim and Ye(2022)}]{kim2022energy}
Beomsu Kim and Jong~Chul Ye. 2022.
\newblock Energy-based contrastive learning of visual representations.
\newblock \emph{Advances in Neural Information Processing Systems}, 35:4358--4369.

\bibitem[{Krithara et~al.(2023)Krithara, Nentidis, Bougiatiotis, and Paliouras}]{krithara2023bioasq}
Anastasia Krithara, Anastasios Nentidis, Konstantinos Bougiatiotis, and Georgios Paliouras. 2023.
\newblock Bioasq-qa: A manually curated corpus for biomedical question answering.
\newblock \emph{Scientific Data}, 10(1):170.

\bibitem[{Latecki et~al.(2007)Latecki, Lazarevic, and Pokrajac}]{latecki2007outlier}
Longin~Jan Latecki, Aleksandar Lazarevic, and Dragoljub Pokrajac. 2007.
\newblock Outlier detection with kernel density functions.
\newblock In \emph{International Workshop on Machine Learning and Data Mining in Pattern Recognition}, pages 61--75. Springer.

\bibitem[{Lee et~al.(2018)Lee, Lee, Lee, and Shin}]{lee2018simple}
Kimin Lee, Kibok Lee, Honglak Lee, and Jinwoo Shin. 2018.
\newblock A simple unified framework for detecting out-of-distribution samples and adversarial attacks.
\newblock \emph{Advances in neural information processing systems}, 31.

\bibitem[{Lewis et~al.(2020)Lewis, Perez, Piktus, Petroni, Karpukhin, Goyal, K{\"u}ttler, Lewis, Yih, Rockt{\"a}schel et~al.}]{lewis2020retrieval}
Patrick Lewis, Ethan Perez, Aleksandra Piktus, Fabio Petroni, Vladimir Karpukhin, Naman Goyal, Heinrich K{\"u}ttler, Mike Lewis, Wen-tau Yih, Tim Rockt{\"a}schel, et~al. 2020.
\newblock Retrieval-augmented generation for knowledge-intensive nlp tasks.
\newblock \emph{Advances in Neural Information Processing Systems}, 33:9459--9474.

\bibitem[{Li et~al.(2020)Li, Zhao, Botta, Ionescu, and Hu}]{li2020copod}
Zheng Li, Yue Zhao, Nicola Botta, Cezar Ionescu, and Xiyang Hu. 2020.
\newblock Copod: copula-based outlier detection.
\newblock In \emph{2020 IEEE international conference on data mining (ICDM)}, pages 1118--1123. IEEE.

\bibitem[{Lin et~al.(2021)Lin, Hilton, and Evans}]{lin2021truthfulqa}
Stephanie Lin, Jacob Hilton, and Owain Evans. 2021.
\newblock Truthfulqa: Measuring how models mimic human falsehoods.
\newblock \emph{arXiv preprint arXiv:2109.07958}.

\bibitem[{Liu et~al.(2020)Liu, Wang, Owens, and Li}]{liu2020energy}
Weitang Liu, Xiaoyun Wang, John Owens, and Yixuan Li. 2020.
\newblock Energy-based out-of-distribution detection.
\newblock \emph{Advances in neural information processing systems}, 33:21464--21475.

\bibitem[{Liu et~al.(2023)Liu, Iter, Xu, Wang, Xu, and Zhu}]{liu2023g}
Yang Liu, Dan Iter, Yichong Xu, Shuohang Wang, Ruochen Xu, and Chenguang Zhu. 2023.
\newblock G-eval: Nlg evaluation using gpt-4 with better human alignment.
\newblock In \emph{The 2023 Conference on Empirical Methods in Natural Language Processing}.

\bibitem[{Manakul et~al.(2023)Manakul, Liusie, and Gales}]{manakul2023selfcheckgpt}
Potsawee Manakul, Adian Liusie, and Mark Gales. 2023.
\newblock Selfcheckgpt: Zero-resource black-box hallucination detection for generative large language models.
\newblock In \emph{The 2023 Conference on Empirical Methods in Natural Language Processing}.

\bibitem[{Ni et~al.(2024)Ni, Bi, Guo, and Cheng}]{ni2024llms}
Shiyu Ni, Keping Bi, Jiafeng Guo, and Xueqi Cheng. 2024.
\newblock When do llms need retrieval augmentation? mitigating llms' overconfidence helps retrieval augmentation.
\newblock \emph{arXiv preprint arXiv:2402.11457}.

\bibitem[{Nussbaum et~al.(2024)Nussbaum, Morris, Duderstadt, and Mulyar}]{nussbaum2024nomic}
Zach Nussbaum, John~X Morris, Brandon Duderstadt, and Andriy Mulyar. 2024.
\newblock Nomic embed: Training a reproducible long context text embedder.
\newblock \emph{arXiv preprint arXiv:2402.01613}.

\bibitem[{Pal et~al.(2022)Pal, Umapathi, and Sankarasubbu}]{pal2022medmcqa}
Ankit Pal, Logesh~Kumar Umapathi, and Malaikannan Sankarasubbu. 2022.
\newblock Medmcqa: A large-scale multi-subject multi-choice dataset for medical domain question answering.
\newblock In \emph{Conference on health, inference, and learning}, pages 248--260. PMLR.

\bibitem[{Ramaswamy et~al.(2000)Ramaswamy, Rastogi, and Shim}]{ramaswamy2000efficient}
Sridhar Ramaswamy, Rajeev Rastogi, and Kyuseok Shim. 2000.
\newblock Efficient algorithms for mining outliers from large data sets.
\newblock In \emph{Proceedings of the 2000 ACM SIGMOD international conference on Management of data}, pages 427--438.

\bibitem[{Ren et~al.(2019)Ren, Liu, Fertig, Snoek, Poplin, Depristo, Dillon, and Lakshminarayanan}]{ren2019likelihood}
Jie Ren, Peter~J Liu, Emily Fertig, Jasper Snoek, Ryan Poplin, Mark Depristo, Joshua Dillon, and Balaji Lakshminarayanan. 2019.
\newblock Likelihood ratios for out-of-distribution detection.
\newblock \emph{Advances in neural information processing systems}, 32.

\bibitem[{Ren et~al.(2022)Ren, Luo, Zhao, Krishna, Saleh, Lakshminarayanan, and Liu}]{ren2022out}
Jie Ren, Jiaming Luo, Yao Zhao, Kundan Krishna, Mohammad Saleh, Balaji Lakshminarayanan, and Peter~J Liu. 2022.
\newblock Out-of-distribution detection and selective generation for conditional language models.
\newblock In \emph{The Eleventh International Conference on Learning Representations}.

\bibitem[{Ren et~al.(2023)Ren, Wang, Qu, Zhao, Liu, Tian, Wu, Wen, and Wang}]{ren2023investigating}
Ruiyang Ren, Yuhao Wang, Yingqi Qu, Wayne~Xin Zhao, Jing Liu, Hao Tian, Hua Wu, Ji-Rong Wen, and Haifeng Wang. 2023.
\newblock Investigating the factual knowledge boundary of large language models with retrieval augmentation.
\newblock \emph{arXiv preprint arXiv:2307.11019}.

\bibitem[{Robertson et~al.(2009)Robertson, Zaragoza et~al.}]{robertson2009probabilistic}
Stephen Robertson, Hugo Zaragoza, et~al. 2009.
\newblock The probabilistic relevance framework: Bm25 and beyond.
\newblock \emph{Foundations and Trends{\textregistered} in Information Retrieval}, 3(4):333--389.

\bibitem[{Sch{\"o}lkopf et~al.(2001)Sch{\"o}lkopf, Platt, Shawe-Taylor, Smola, and Williamson}]{scholkopf2001estimating}
Bernhard Sch{\"o}lkopf, John~C Platt, John Shawe-Taylor, Alex~J Smola, and Robert~C Williamson. 2001.
\newblock Estimating the support of a high-dimensional distribution.
\newblock \emph{Neural computation}, 13(7):1443--1471.

\bibitem[{Shi et~al.(2023)Shi, Chen, Misra, Scales, Dohan, Chi, Sch{\"a}rli, and Zhou}]{shi2023large}
Freda Shi, Xinyun Chen, Kanishka Misra, Nathan Scales, David Dohan, Ed~H Chi, Nathanael Sch{\"a}rli, and Denny Zhou. 2023.
\newblock Large language models can be easily distracted by irrelevant context.
\newblock In \emph{International Conference on Machine Learning}, pages 31210--31227. PMLR.

\bibitem[{Shuster et~al.(2021)Shuster, Poff, Chen, Kiela, and Weston}]{shuster2021retrieval}
Kurt Shuster, Spencer Poff, Moya Chen, Douwe Kiela, and Jason Weston. 2021.
\newblock Retrieval augmentation reduces hallucination in conversation.
\newblock In \emph{Findings of the Association for Computational Linguistics: EMNLP 2021}, pages 3784--3803.

\bibitem[{Simes(1986)}]{simes1986improved}
R~John Simes. 1986.
\newblock An improved bonferroni procedure for multiple tests of significance.
\newblock \emph{Biometrika}, 73(3):751--754.

\bibitem[{Spithourakis and Riedel(2018)}]{spithourakis2018numeracy}
Georgios Spithourakis and Sebastian Riedel. 2018.
\newblock Numeracy for language models: Evaluating and improving their ability to predict numbers.
\newblock In \emph{Proceedings of the 56th Annual Meeting of the Association for Computational Linguistics (Volume 1: Long Papers)}, pages 2104--2115.

\bibitem[{Sun et~al.(2022)Sun, Ming, Zhu, and Li}]{sun2022out}
Yiyou Sun, Yifei Ming, Xiaojin Zhu, and Yixuan Li. 2022.
\newblock Out-of-distribution detection with deep nearest neighbors.
\newblock In \emph{International Conference on Machine Learning}, pages 20827--20840. PMLR.

\bibitem[{Talmor et~al.(2018)Talmor, Herzig, Lourie, and Berant}]{talmor2018commonsenseqa}
Alon Talmor, Jonathan Herzig, Nicholas Lourie, and Jonathan Berant. 2018.
\newblock Commonsenseqa: A question answering challenge targeting commonsense knowledge.
\newblock \emph{arXiv preprint arXiv:1811.00937}.

\bibitem[{Tan et~al.(2022)Tan, Gu, Tao, Ling, Xu, Hu, Geng, and Jiang}]{tan2022tegtok}
Chao-Hong Tan, Jia-Chen Gu, Chongyang Tao, Zhen-Hua Ling, Can Xu, Huang Hu, Xiubo Geng, and Daxin Jiang. 2022.
\newblock Tegtok: Augmenting text generation via task-specific and open-world knowledge.
\newblock In \emph{Findings of the Association for Computational Linguistics: ACL 2022}, pages 1597--1609.

\bibitem[{Wang et~al.(2023)Wang, Li, Sun, and Liu}]{wang2023self}
Yile Wang, Peng Li, Maosong Sun, and Yang Liu. 2023.
\newblock Self-knowledge guided retrieval augmentation for large language models.
\newblock \emph{arXiv preprint arXiv:2310.05002}.

\bibitem[{Wornow et~al.(2023)Wornow, Xu, Thapa, Patel, Steinberg, Fleming, Pfeffer, Fries, and Shah}]{wornow2023shaky}
Michael Wornow, Yizhe Xu, Rahul Thapa, Birju Patel, Ethan Steinberg, Scott Fleming, Michael~A Pfeffer, Jason Fries, and Nigam~H Shah. 2023.
\newblock The shaky foundations of large language models and foundation models for electronic health records.
\newblock \emph{npj Digital Medicine}, 6(1):135.

\bibitem[{Xiao et~al.(2023)Xiao, Liu, Zhang, and Muennighof}]{xiao2023c}
Shitao Xiao, Zheng Liu, Peitian Zhang, and Niklas Muennighof. 2023.
\newblock C-pack: Packaged resources to advance general chinese embedding.
\newblock \emph{arXiv preprint arXiv:2309.07597}.

\bibitem[{Xiong et~al.(2024)Xiong, Jin, Lu, and Zhang}]{xiong2024benchmarking}
Guangzhi Xiong, Qiao Jin, Zhiyong Lu, and Aidong Zhang. 2024.
\newblock Benchmarking retrieval-augmented generation for medicine.
\newblock \emph{arXiv preprint arXiv:2402.13178}.

\bibitem[{Yan et~al.(2024)Yan, Gu, Zhu, and Ling}]{yan2024corrective}
Shi-Qi Yan, Jia-Chen Gu, Yun Zhu, and Zhen-Hua Ling. 2024.
\newblock Corrective retrieval augmented generation.
\newblock \emph{arXiv preprint arXiv:2401.15884}.

\bibitem[{Yang et~al.(2021)Yang, Zhou, Li, and Liu}]{yang2021generalized}
Jingkang Yang, Kaiyang Zhou, Yixuan Li, and Ziwei Liu. 2021.
\newblock Generalized out-of-distribution detection: A survey.
\newblock \emph{arXiv preprint arXiv:2110.11334}.

\bibitem[{Yang et~al.(2015)Yang, Yih, and Meek}]{yang2015wikiqa}
Yi~Yang, Wen-tau Yih, and Christopher Meek. 2015.
\newblock Wikiqa: A challenge dataset for open-domain question answering.
\newblock In \emph{Proceedings of the 2015 conference on empirical methods in natural language processing}, pages 2013--2018.

\bibitem[{Yang et~al.(2023)Yang, Chern, Qiu, Neubig, and Liu}]{yang2023alignment}
Yuqing Yang, Ethan Chern, Xipeng Qiu, Graham Neubig, and Pengfei Liu. 2023.
\newblock Alignment for honesty.
\newblock \emph{arXiv preprint arXiv:2312.07000}.

\bibitem[{Yin et~al.(2023)Yin, Sun, Guo, Wu, Qiu, and Huang}]{yin2023large}
Zhangyue Yin, Qiushi Sun, Qipeng Guo, Jiawen Wu, Xipeng Qiu, and Xuanjing Huang. 2023.
\newblock Do large language models know what they don't know?
\newblock \emph{arXiv preprint arXiv:2305.18153}.

\bibitem[{Yu et~al.(2022)Yu, Iter, Wang, Xu, Ju, Sanyal, Zhu, Zeng, and Jiang}]{yu2022generate}
Wenhao Yu, Dan Iter, Shuohang Wang, Yichong Xu, Mingxuan Ju, Soumya Sanyal, Chenguang Zhu, Michael Zeng, and Meng Jiang. 2022.
\newblock Generate rather than retrieve: Large language models are strong context generators.
\newblock In \emph{The Eleventh International Conference on Learning Representations}.

\bibitem[{Zhao et~al.(2019)Zhao, Nasrullah, and Li}]{zhao2019pyod}
Yue Zhao, Zain Nasrullah, and Zheng Li. 2019.
\newblock \href {http://jmlr.org/papers/v20/19-011.html} {Pyod: A python toolbox for scalable outlier detection}.
\newblock \emph{Journal of Machine Learning Research}, 20(96):1--7.

\bibitem[{Zhou et~al.(2021)Zhou, Liu, and Chen}]{zhou2021contrastive}
Wenxuan Zhou, Fangyu Liu, and Muhao Chen. 2021.
\newblock Contrastive out-of-distribution detection for pretrained transformers.
\newblock In \emph{Proceedings of the 2021 Conference on Empirical Methods in Natural Language Processing}, pages 1100--1111.

\end{thebibliography}

\appendix

\section{Appendix}
\label{sec:appendix}

\subsection{Implementation Details}

\paragraph{Retrievers / Embedding Models.}
(1) \textit{BM25} \cite{robertson2009probabilistic} is a bag-of-word based retriever function.
(2) \textit{BGE} \cite{xiao2023c} is a general embedding model that maps the input text to a low-dimensional dense vector. We use the \texttt{bge-base-en-v1.5} version, a model trained with cosine similarity and a dimension size of $768$.
(3) \textit{Contriever} \cite{izacardunsupervised} is a contrastive dense retriever model developed by Facebook.
(4) \textit{MedCPT} \cite{jin2023medcpt} is a contrastive transformer-based embedding model trained on PubMed search logs.
(5) \textit{Nomic} \cite{nussbaum2024nomic} is a long context-length text encoder. We adopt the \texttt{nomic-embed-text-v1-unsupervised} version, which uses dot product as the similarity metric and has an embedding size of $768$.
(6) \textit{Jina} \cite{gunther2023jina} is a long context-length embedding model based on BERT architecture. We use the \texttt{jina-embeddings-v2-base-en} version that uses dot product as the similarity metric and has an embedding size of $768$.
We implement the RAG system using FAISS index \cite{johnson2019billion} with maximum inner product search (MIPS) algorithms.

\paragraph{Synthetic Query Generation.} Synthetic queries are created by sampling document chunks from the corpus and using them as context to prompt \texttt{GPT-3.5-turbo}. The prompt for generating synthetic queries and examples of generated queries are included in Table~\ref{tab:synthetic_prompt} and Table~\ref{tab:synthetic_query}, respectively.

\paragraph{Outlier Detection-based Baselines.} We apply outlier detection-based algorithms to model the density of the queries' embedding distribution and then flag those with their embeddings falling into the tail of this distribution as OoK queries. To ensure a fair comparison with the training-free methods, we train the outlier detection models using $300$ synthetic IK queries and report the AUROC and AUPRC on a balanced testing set with $300$ IK queries and $300$ OoK queries. We use the PyOD package \cite{zhao2019pyod} with default parameters for implementing these outlier detection baselines. 

\paragraph{LM-based Relevance Score.} We use \texttt{GPT-3.5-turbo} and \texttt{GPT-4-turbo-preview} models accessed through Microsoft Azure as LM relevance evaluators. The LM evaluator is asked to reason in a step-by-step manner and provide text feedback along with a numerical score between $0$ and $1$ with a higher score indicating better relevance.
Long document chunks are truncated to fit the context length.
The prompt for generating relevance score is included in Table~\ref{tab:lm_score_prompt}, which is modified from the default prompt template used by the \texttt{ContextRelevancyEvaluator}\footnote{\url{https://docs.llamaindex.ai/en/stable/api_reference/evaluation/context_relevancy/}} in the Llama-index framework.
To analyze the output of the LM-based relevance score, we additionally ask LM to provide a summary of feedback on the results.

\subsection{Discussions}

\paragraph{Risk of OoK Queries.} As OoK queries are beyond the context of the corpus, answering these queries with retrieved documents is likely to increase the risk of generating hallucinated responses. For instance, using the chain-of-thought prompting strategy, GPT-3.5 answers PubMedQA and CommonsenseQA questions with an accuracy of $48.6\%$ and $59.3\%$, respectively. In contrast, retrieving from the Textbooks corpus reduces the accuracy to $22.9\%$ and $29.0\%$.
To improve the generation reliability, there are several potential strategies to deal with an OoK query once it has been detected, including: (1) \textit{Rejection}: refusing to respond to the query; (2) \textit{Direct Generation}: skipping the retrieval process and invoking the LM to directly answer the query; (3) \textit{Broader Search}: searching with a broader knowledge base, such as the Internet, to answer the query; and (4) \textit{Human Intervention}: triggering an alert to request for assistance from human experts. The appropriate strategy should be decided based on the specific application scenario.

\paragraph{Synthetic Query Quality.} Currently, we generate synthetic queries by asking an LM to produce questions based on individual document chunks. However, such method might produce overly simple questions that only directly relate to a single document chunk. Consequently, the distribution of the synthetic queries could deviate from that of the real IK queries, which tend to be more complex in nature, involving knowledge across multiple chunks of the corpus. This complexity is evident in the queries found in the Textbooks corpus and the MedQA-US dataset, as illustrated in Figure~\ref{fig:hist_all}.
To generate more challenging queries that require a general understanding of the subject in the corpus, one straightforward way is to provide the LM with the entire corpus and ask it to generate synthetic queries accordingly. However, this contradicts the motivation of deploying RAG and is generally not feasible due to the limited context window of LMs. 
Instead, one can employ context-aware chunking strategies, such as hierarchical or semantic chunking, to generate more complex synthetic queries that involve knowledge from multiple fixed-size chunks. 
We leave this for future exploration.

\subsection{Additional Results}

\paragraph{Impact of Different Sample Sizes.} Our main experiments are conducted by drawing 300 I.I.D. samples from the In-Knowledge query distribution. To study the impact of different sample sizes and further validate the sampling process, we include additional experiments with 100 and 500 samples using energy score as the test statistic and summarize the results in Table~\ref{tab:sample_size} (averaged over 5 independent runs). We observe that the performance is consistent across different sample sizes. In particular, the proposed method can achieve high AUROC and AUPRC using only 100 samples for the estimation of the query distribution.

\begin{table}
\centering
\caption{Results on Textbooks corpus with different sample sizes.}
\label{tab:sample_size}
\begin{subtable}{\linewidth}
\centering
\caption{AUROC}
\setlength{\extrarowheight}{0pt}
\addtolength{\extrarowheight}{\aboverulesep}
\addtolength{\extrarowheight}{\belowrulesep}
\setlength{\aboverulesep}{0pt}
\setlength{\belowrulesep}{0pt}
\resizebox{0.98\linewidth}{!}{
\begin{tabular}{c|c|c|c}
\toprule
           & 100 samples & 300 samples & 500 samples \\ \midrule
MMLU-US    & 0.8585      & 0.8611      & 0.8562      \\ 
PubMedQA   & 0.9529      & 0.9520      & 0.9548      \\ 
TruthfulQA & 0.9999      & 0.9998      & 0.9998      \\ \bottomrule
\end{tabular}
}
\vspace{2mm}
\end{subtable}

\begin{subtable}{\linewidth}
\centering
\caption{AUPRC}
\setlength{\extrarowheight}{0pt}
\addtolength{\extrarowheight}{\aboverulesep}
\addtolength{\extrarowheight}{\belowrulesep}
\setlength{\aboverulesep}{0pt}
\setlength{\belowrulesep}{0pt}
\resizebox{0.98\linewidth}{!}{
\begin{tabular}{c|c|c|c}
\toprule
           & 100 samples & 300 samples & 500 samples \\ \midrule
MMLU-US    & 0.8752      & 0.8716      & 0.8699      \\ 
PubMedQA   & 0.9425      & 0.9375      & 0.9408      \\ 
TruthfulQA & 0.9999      & 0.9998      & 0.9998      \\ \bottomrule
\end{tabular}
}
\end{subtable}
\end{table}

\paragraph{Results on Entire Datasets.} Most of our previous experiments are conducted on a sampled subset as the considered list of datasets has varying sizes (e.g., PubMedQA has 500 questions whereas CommonsenseQA has 12247 questions). Here we conduct additional experiments on the entire dataset of MMLU-US, BioASQ-Y/N, PubMedQA, and TruthfulQA, and summarize the results in Table~\ref{tab:res_entire_dataset}. We observe that these results are consistent with those produced from sampled subsets.

\begin{table}
\centering
\caption{Results of experiments with entire datasets on Textbooks corpus.}
\label{tab:res_entire_dataset}
\setlength{\extrarowheight}{0pt}
\addtolength{\extrarowheight}{\aboverulesep}
\addtolength{\extrarowheight}{\belowrulesep}
\setlength{\aboverulesep}{0pt}
\setlength{\belowrulesep}{0pt}
\resizebox{\linewidth}{!}{
\begin{tabular}{c|c|c|c|c}
\toprule
      & MMLU-US & BioASQ-Y/N & PubMedQA & TruthfulQA \\ \midrule
AUROC & 0.8585  & 0.9699     & 0.9529   & 0.9999     \\
AUPRC & 0.8752  & 0.9607     & 0.9425   & 0.9999 \\ \bottomrule
\end{tabular}
}
\end{table}

\begin{sidewaysfigure*}
    \includegraphics[width=\linewidth]{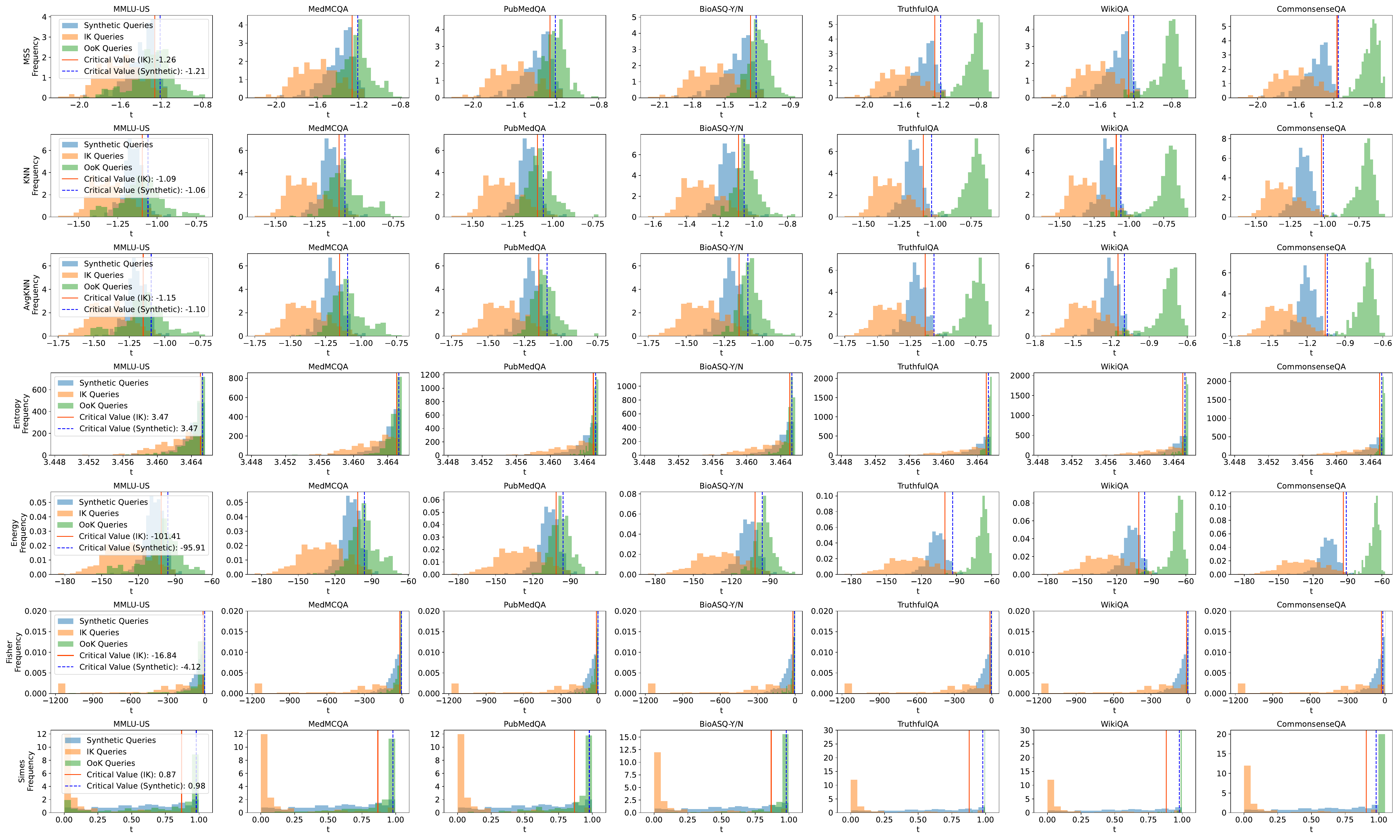}
    \caption{Histograms of different test statistics on the Textbooks corpus with Contriever as the embedding model.}
    \label{fig:hist_all}
\end{sidewaysfigure*}

\begin{sidewaysfigure*}
    \includegraphics[width=\linewidth]{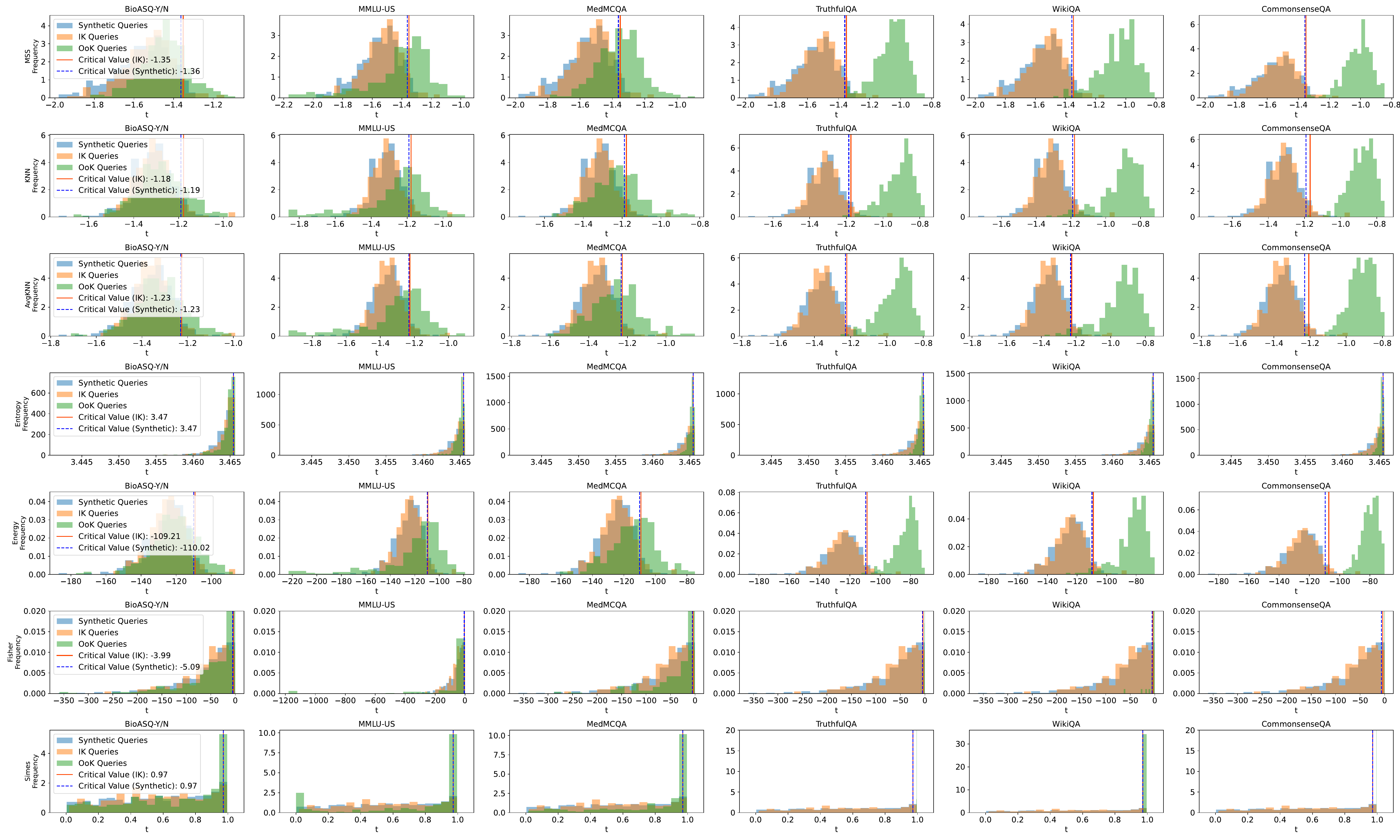}
    \caption{Histograms of different test statistics on the PubMed corpus with Contriever as the embedding model.}
    \label{fig:hist_all_pubmed}
\end{sidewaysfigure*}

\begin{figure*}
    \centering
    \includegraphics[width=0.96\textwidth]{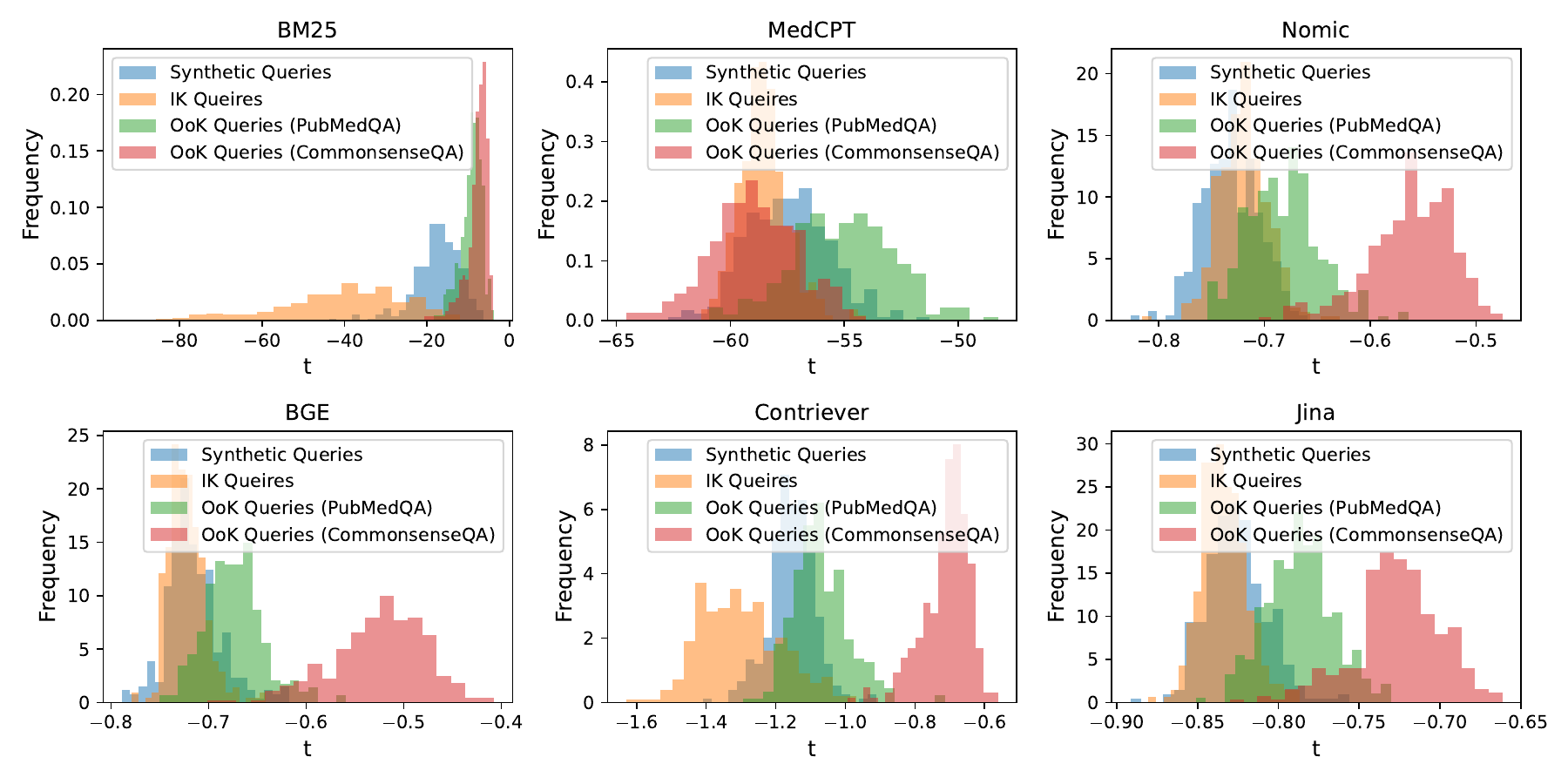}
    \vspace{-4mm}
    \caption{Histograms with different embedding models on the Textbooks corpus.}
    \label{fig:retriever_hist}
\end{figure*}

\paragraph{Extended Results from Main Paper.}
Table~\ref{tab:res_auprc} presents the AUPRC results of different test statistics. Table~\ref{tab:synthetic_extended} and Table~\ref{tab:synthetic_extended_pubmed} present the extended results with synthetic queries on the Textbooks and the PubMed corpora, respectively. Additionally, we show the histograms of different test statistics with Contriever on both corpora in Figure~\ref{fig:hist_all} and Figure~\ref{fig:hist_all_pubmed} respectively. Finally, Figure~\ref{fig:retriever_hist} compares the histograms produced with different embedding models on the Textbooks corpus.

\begin{table*}
\centering
\caption{AUPRC results of different test statistics.}
\label{tab:res_auprc}
\vspace{1mm}
\setlength{\extrarowheight}{0pt}
\addtolength{\extrarowheight}{\aboverulesep}
\addtolength{\extrarowheight}{\belowrulesep}
\setlength{\aboverulesep}{0pt}
\setlength{\belowrulesep}{0pt}
\resizebox{0.75\linewidth}{!}{
\begin{tabular}{c|c|ccccccc} 
\toprule
\multirow{2}{*}{\textbf{Corpus}}    & \multirow{2}{*}{\textbf{Dataset}}                          & \multicolumn{7}{c}{\textbf{Test Statistics}}                            \\ 
\cline{3-9}
                           &                                                   & MSS    & KNN    & AvgKNN & Entropy & Energy & Fisher & Simes   \\ 
\midrule
\multirow{7}{*}{Textbooks} & MMLU-US                                           & 0.8676 & 0.8702 & 0.8740 & 0.7828  & 0.8741 & 0.8730 & 0.8737  \\
                           & MedMCQA                                           & 0.9405 & 0.9379 & 0.9434 & 0.8599  & 0.9436 & 0.9423 & 0.9434  \\
                           & {\cellcolor[rgb]{0.94,0.94,0.94}}PubMedQA         & 0.9525 & 0.9331 & 0.9454 & 0.9146  & 0.9458 & 0.9444 & 0.9495  \\
                           & {\cellcolor[rgb]{0.94,0.94,0.94}}BioASQ-Y/N       & 0.9611 & 0.9609 & 0.9659 & 0.8928  & 0.9660 & 0.9649 & 0.9693  \\
                           & {\cellcolor[rgb]{0.86,0.86,0.86}}TruthfulQA       & 0.9998 & 0.9999 & 0.9999 & 0.9629  & 0.9999 & 0.9994 & 0.9987  \\
                           & {\cellcolor[rgb]{0.86,0.86,0.86}}WikiQA           & 0.9982 & 0.9986 & 0.9986 & 0.9639  & 0.9986 & 0.9981 & 0.9976  \\
                           & {\cellcolor[rgb]{0.86,0.86,0.86}}CommonsenseQA    & 0.9999 & 0.9999 & 0.9999 & 0.9668  & 0.9999 & 0.9994 & 0.9988  \\
\midrule
\multirow{7}{*}{PubMed}    & BioASQ-Y/N                                     & 0.6544 & 0.6131 & 0.6006 & 0.5713  & 0.6010 & 0.6011 & 0.6214  \\
                           & {\cellcolor[rgb]{0.94,0.94,0.94}}MMLU-US       & 0.8013 & 0.7430 & 0.7516 & 0.6400  & 0.7527 & 0.7578 & 0.7668  \\
                           & {\cellcolor[rgb]{0.94,0.94,0.94}}MedMCQA       & 0.8013 & 0.7573 & 0.7672 & 0.6973  & 0.7687 & 0.7718 & 0.7845  \\
                           & {\cellcolor[rgb]{0.86,0.86,0.86}}TruthfulQA    & 0.8388 & 0.9970 & 0.9974 & 0.6454  & 0.9975 & 0.9958 & 0.9922  \\
                           & {\cellcolor[rgb]{0.86,0.86,0.86}}WikiQA        & 0.9980 & 0.9921 & 0.9926 & 0.7061  & 0.9927 & 0.9907 & 0.9874  \\
                           & {\cellcolor[rgb]{0.86,0.86,0.86}}CommonsenseQA & 0.9946 & 0.9984 & 0.9989 & 0.7012  & 0.9989 & 0.9968 & 0.9929  \\
\bottomrule
\end{tabular}
}
\end{table*}

\begin{table*}
\centering
\caption{Extended results with synthetic queries on the Textbooks corpus.}
\label{tab:synthetic_extended}
\vspace{1mm}
\setlength{\extrarowheight}{0pt}
\addtolength{\extrarowheight}{\aboverulesep}
\addtolength{\extrarowheight}{\belowrulesep}
\setlength{\aboverulesep}{0pt}
\setlength{\belowrulesep}{0pt}
\resizebox{0.75\linewidth}{!}{
\begin{tabular}{c|c|c|ccccccc} 
\toprule
\multirow{2}{*}{\textbf{Dataset}}    & \multirow{2}{*}{\textbf{Data Source}}    & \multirow{2}{*}{\begin{tabular}[c]{@{}l@{}}\textbf{Metric}\\$\alpha=5\%$\end{tabular}}                          & \multicolumn{7}{c}{\textbf{Test Statistics}}                            \\ 
\cline{4-10}
                        &                                       &           & MSS    & KNN    & AvgKNN & Entropy & Energy & Fisher & Simes   \\ 
\midrule
\multirow{4}{*}{MMLU-US} & \multirow{2}{*}{In-knowledge Queries} & TPR       & 0.4996 & 0.5180 & 0.5393 & 0.3406  & 0.5386 & 0.5406 & 0.5388 \\
                        &                                        & DER       & 0.2763 & 0.2676 & 0.2558 & 0.3561  & 0.2570 & 0.2560 & 0.2569  \\
\cline{2-10}
                        & \multirow{2}{*}{Synthetic Queries}    & TPR       & 0.3666 & 0.3566 & 0.3413 & 0.1336  & 0.3403 & 0.3440 & 0.3200  \\
                        &                                       & DER       & 0.3300 & 0.3331 & 0.3393 & 0.4410  & 0.3400 & 0.3375 & 0.3495  \\
\midrule
\multirow{4}{*}{MedMCQA} & \multirow{2}{*}{In-knowledge Queries} & TPR      & 0.6880 & 0.6650 & 0.7023 & 0.4530  & 0.7023 & 0.7010 & 0.7015 \\
                        &                                        & DER      & 0.1825 & 0.1936 & 0.1745 & 0.3000  & 0.1753 & 0.1761 & 0.1761  \\
\cline{2-10}
                        & \multirow{2}{*}{Synthetic Queries}    & TPR       & 0.5060 & 0.4730 & 0.4766 & 0.2350  & 0.4776 & 0.4780 & 0.4633  \\
                        &                                       & DER       & 0.2601 & 0.2753 & 0.2714 & 0.3906  & 0.2710 & 0.2706 & 0.2781  \\
\midrule
\multirow{4}{*}{PubMedQA} & \multirow{2}{*}{In-knowledge Queries} & TPR     & 0.7620 & 0.6523 & 0.7250 & 0.6026  & 0.7283 & 0.7233 & 0.7463 \\
                        &                                       & DER       & 0.1451 & 0.2003 & 0.1636 & 0.2251  & 0.1623 & 0.1649 & 0.1533  \\
\cline{2-10}
                        & \multirow{2}{*}{Synthetic Queries}    & TPR       & 0.5333 & 0.4116 & 0.4353 & 0.3643  & 0.4353 & 0.4380 & 0.4429  \\
                        &                                       & DER       & 0.2465 & 0.3055 & 0.2923 & 0.3264  & 0.2923 & 0.2906 & 0.2893  \\
\midrule
\multirow{4}{*}{BioASQ-Y/N} & \multirow{2}{*}{In-knowledge Queries} & TPR   & 0.8073 & 0.8303 & 0.8639 & 0.5836  & 0.8653 & 0.8630 & 0.8610 \\
                        &                                       & DER       & 0.1230 & 0.1110 & 0.0941 & 0.2345  & 0.0938 & 0.0948 & 0.0956  \\
\cline{2-10}
                        & \multirow{2}{*}{Synthetic Queries}    & TPR       & 0.6103 & 0.5823 & 0.5943 & 0.3150  & 0.5933 & 0.5940 & 0.5730  \\
                        &                                       & DER       & 0.2081 & 0.2210 & 0.2126 & 0.3506  & 0.2131 & 0.2125 & 0.2250  \\
\midrule
\multirow{4}{*}{TruthfulQA} & \multirow{2}{*}{In-knowledge Queries} & TPR   & 0.9993 & 1.0    & 1.0    & 0.8160  & 1.0    & 1.0    & 0.9996 \\
                        &                                       & DER       & 0.0153 & 0.0088 & 0.0101 & 0.1186  & 0.0116 & 0.0471 & 0.0241  \\
\cline{2-10}
                        & \multirow{2}{*}{Synthetic Queries}    & TPR       & 0.9990 & 0.9990 & 0.9990 & 0.6446  & 0.9990 & 0.9990 & 0.9990  \\
                        &                                       & DER       & 0.0081 & 0.0025 & 0.0035 & 0.1861  & 0.0039 & 0.0181 & 0.0069  \\
\midrule
\multirow{4}{*}{WikiQA} & \multirow{2}{*}{In-knowledge Queries} & TPR       & 0.9869 & 0.9906 & 0.9916 & 0.8260  & 0.9916 & 0.9916 & 0.9913 \\
                        &                                       & DER       & 0.0319 & 0.0264 & 0.0253 & 0.1126  & 0.0254 & 0.0301 & 0.0286  \\
\cline{2-10}
                        & \multirow{2}{*}{Synthetic Queries}    & TPR       & 0.9760 & 0.9843 & 0.9816 & 0.6500  & 0.9816 & 0.9816 & 0.9773  \\
                        &                                       & DER       & 0.0238 & 0.0181 & 0.0185 & 0.1831  & 0.0185 & 0.0184 & 0.0181  \\
\midrule
\multirow{4}{*}{CommonsenseQA} & \multirow{2}{*}{In-knowledge Queries} & TPR & 1.0    & 1.0    & 1.0    & 0.8420  & 1.0    & 1.0    & 1.0 \\
                        &                                       & DER        & 0.0113 & 0.0035 & 0.0040 & 0.1053  & 0.0065 & 0.0538 & 0.0239  \\
\cline{2-10}
                        & \multirow{2}{*}{Synthetic Queries}    & TPR       & 0.9996 & 1.0    & 1.0    & 0.6253  & 1.0    & 1.0    & 1.0  \\
                        &                                       & DER       & 0.0070 & 0.0023 & 0.0021 & 0.1951  & 0.0023 & 0.0221 & 0.0064  \\
\bottomrule
\end{tabular}
}
\end{table*}

\begin{table*}
\centering
\caption{Extended results with synthetic queries on the PubMed corpus.}
\label{tab:synthetic_extended_pubmed}
\vspace{1mm}
\setlength{\extrarowheight}{0pt}
\addtolength{\extrarowheight}{\aboverulesep}
\addtolength{\extrarowheight}{\belowrulesep}
\setlength{\aboverulesep}{0pt}
\setlength{\belowrulesep}{0pt}
\resizebox{0.9\linewidth}{!}{
\begin{tabular}{c|c|c|ccccccc} 
\toprule
\multirow{2}{*}{\textbf{Dataset}}    & \multirow{2}{*}{\textbf{Data Source}}    & \multirow{2}{*}{\begin{tabular}[c]{@{}l@{}}\textbf{Metric}\\$\alpha=5\%$\end{tabular}}                          & \multicolumn{7}{c}{\textbf{Test Statistics}}                            \\ 
\cline{4-10}
                        &                                       &           & MSS    & KNN    & AvgKNN & Entropy & Energy & Fisher & Simes   \\ 
\midrule
\multirow{4}{*}{BioASQ-Y/N} & \multirow{2}{*}{In-knowledge Queries} & TPR   & 0.1743 & 0.1566 & 0.1500 & 0.0676  & 0.1516 & 0.1460 & 0.1616 \\
                        &                                       & DER       & 0.4376 & 0.4455 & 0.4514 & 0.4916  & 0.4506 & 0.4530 & 0.4445  \\
\cline{2-10}
                        & \multirow{2}{*}{Synthetic Queries}    & TPR       & 0.1786 & 0.1743 & 0.1590 & 0.1026  & 0.1613 & 0.1586 & 0.1730  \\
                        &                                       & DER       & 0.4389 & 0.4461 & 0.4506 & 0.4816  & 0.4493 & 0.4504 & 0.4406  \\
\midrule
\multirow{4}{*}{MMLU-US} & \multirow{2}{*}{In-knowledge Queries} & TPR       & 0.4793 & 0.3706 & 0.4036 & 0.1326  & 0.4073 & 0.4156 & 0.4246 \\
                        &                                        & DER       & 0.2858 & 0.3394 & 0.3250 & 0.4599  & 0.3230 & 0.3183 & 0.3119  \\
\cline{2-10}
                        & \multirow{2}{*}{Synthetic Queries}    & TPR       & 0.4856 & 0.3976 & 0.4160 & 0.1639  & 0.4189 & 0.4283 & 0.4403  \\
                        &                                       & DER       & 0.2856 & 0.3346 & 0.3219 & 0.4509  & 0.3203 & 0.3151 & 0.3068  \\
\midrule
\multirow{4}{*}{MedMCQA} & \multirow{2}{*}{In-knowledge Queries} & TPR      & 0.4793 & 0.3236 & 0.3586 & 0.1790  & 0.3633 & 0.3703 & 0.3933 \\
                        &                                        & DER      & 0.2865 & 0.3625 & 0.3471 & 0.4366  & 0.3448 & 0.3411 & 0.3283  \\
\cline{2-10}
                        & \multirow{2}{*}{Synthetic Queries}    & TPR       & 0.4873 & 0.3486 & 0.3673 & 0.2336  & 0.3700 & 0.3800 & 0.4053  \\
                        &                                       & DER       & 0.2848 & 0.3586 & 0.3465 & 0.4163  & 0.3450 & 0.3396 & 0.3241  \\
\midrule
\multirow{4}{*}{TruthfulQA} & \multirow{2}{*}{In-knowledge Queries} & TPR   & 0.9960 & 0.9960 & 0.9966 & 0.0930  & 0.9973 & 0.9976 & 0.9956 \\
                        &                                       & DER       & 0.0251 & 0.0213 & 0.0244 & 0.4786  & 0.0253 & 0.0321 & 0.0249  \\
\cline{2-10}
                        & \multirow{2}{*}{Synthetic Queries}    & TPR       & 0.9963 & 0.9973 & 0.9973 & 0.1410  & 0.9976 & 0.9976 & 0.9960  \\
                        &                                       & DER       & 0.0273 & 0.0268 & 0.0278 & 0.4623  & 0.0286 & 0.0441 & 0.0265  \\
\midrule
\multirow{4}{*}{WikiQA} & \multirow{2}{*}{In-knowledge Queries} & TPR       & 0.9780 & 0.9653 & 0.9680 & 0.1900  & 0.9683 & 0.9690 & 0.9726 \\
                        &                                       & DER       & 0.0335 & 0.0426 & 0.0413 & 0.4309  & 0.0413 & 0.0408 & 0.0361  \\
\cline{2-10}
                        & \multirow{2}{*}{Synthetic Queries}    & TPR       & 0.9963 & 0.9666 & 0.9703 & 0.2376  & 0.9713 & 0.9730 & 0.9750  \\
                        &                                       & DER       & 0.0273 & 0.0469 & 0.0443 & 0.4140  & 0.0441 & 0.0418 & 0.0381  \\
\midrule
\multirow{4}{*}{CommonsenseQA} & \multirow{2}{*}{In-knowledge Queries} & TPR & 0.9990 & 0.9989 & 0.9989 & 0.1620  & 0.9989 & 0.9993 & 0.9990 \\
                        &                                       & DER        & 0.0155 & 0.0138 & 0.0135 & 0.4450  & 0.0141 & 0.0539 & 0.0231  \\
\cline{2-10}
                        & \multirow{2}{*}{Synthetic Queries}    & TPR       & 0.9990 & 0.9993 & 0.9996 & 0.2233  & 0.9996 & 0.9996 & 0.9993  \\
                        &                                       & DER       & 0.0188 & 0.0188 & 0.0176 & 0.4211  & 0.0193 & 0.0798 & 0.0249  \\
\bottomrule
\end{tabular}
}
\end{table*}

\subsection{Prompts and Examples of Queries and Responses}
Table~\ref{tab:synthetic_prompt} presents the prompt for generating the synthetic queries. Table~\ref{tab:lm_score_prompt} presents the prompt used for generating the LM-based relevance score. Table~\ref{tab:hallucinated_LM_score} shows examples of hallucinated responses from the LM-based relevance evaluator.

\begin{table*}
\centering
\caption{Prompt for generating synthetic queries.}
\label{tab:synthetic_prompt}
\begin{tabular}{p{12cm}}
\toprule
\textbf{System}\\
\midrule
You are a professor setting up quiz questions for medical students. \\
The qustions should be based only on context from textbook and should be diverse in nature. \\
Below are some sample questions. \\
\\
\{\textit{Examples}\}\\
\\
\midrule
\textbf{User} \\
\midrule
Below is a chunk of context from textbook. \\
\\
---\\
\{\textit{Context}\}\\
---\\
\\
Given the context information, please generate similar question following the json format.\\
\bottomrule
\end{tabular}
\end{table*}

\begin{table*}
\centering
\caption{Prompt for LM-based relevance score.}
\label{tab:lm_score_prompt}
\begin{tabular}{p{12cm}}
\toprule
\textbf{System}\\
\midrule
Your task is to evaluate if the query is relevant to the retrieved context from the document sources.\\
Return the evaluation result by strictly following the following json format:\\
\\
---\\
\{\\
    "feedback": $[FEEDBACK]$,\\
    "score": $[SCORE]$,\\
\}\\
---\\
\\
$[FEEDBACK]$ should be a string with the detailed feedback with step-by-step explanations on the response.\\
$[SCORE]$ should be a float number between 0 and 1, representing the relevance score assigned to the query.\\
0 represents no relevance (the query is from a different subject of matter) and 1 represents perfect relevance (query is highly related to the context).\\
Partial scores are allowed and encouraged.\\
\midrule
\textbf{User} \\
\midrule
Query:\\
\\
\{\textit{query}\}\\
\\
Context:\\
\\
\{\textit{context}\}\\
\\
Return the result in json format. Do not return any other information other than the json object.\\
\bottomrule
\end{tabular}
\end{table*}

\begin{table*}
\centering
\caption{Examples of hallucinated response from LM-based relevance evaluator.}
\label{tab:hallucinated_LM_score}
\footnotesize
\label{tab:lm_score_prompt}
\begin{subtable}{\textwidth}
\centering
\caption{GPT-3.5}\label{tab:gpt3.5_example}
\begin{tabular}{p{12cm}}
\toprule
\textbf{GPT-3.5 Example 1}\\
\midrule
\textbf{Query}: When you play games with children you're doing what to them?\\
\textbf{Document}: Intervention is needed for caregivers who disclose IPV. It is appropriate to show concern and to provide available community resources. It is important to assess for the safety of the victim and the children. In some states, physicians are mandated to report IPV. Information for families that provides details about community resources and state laws is helpful. Youth violence is a leading cause of pediatric mortality in the United States. Homicide is the second leading cause of death for all children 1 to 19 years of age. Each year nearly 6000 children, primarily adolescents, are victims of homicide, and 4000 How are things at home and at school? Who lives with you? How do you get along with your family members? What do you like to do with them? What do you do if something is bothering you? Do you feel safe at home? Do people fight at home? What do they fight about? How do they fight? Do your friends get into fights often? How about you? When was your last physical fight? Have you ever been injured during a fight? Has anyone you know been injured or killed? Have you ever been forced to have sex against your will? Have you ever been threatened with a gun or a knife? How do you avoid getting in fights? Do you carry a weapon for self-defense? FOR THE PARENT Do you have any concerns about your child? Who helps with your children? How do you feel about your neighborhood? Do you feel safe at home? Is there any fighting or violence at home? Does anyone at home use drugs? \ldots\\
\textbf{Output}: \{'score': 1, 'feedback': "The query is highly relevant to the context. The context discusses the importance of playing developmentally appropriate games with children for their cognitive development and school performance. The query specifically asks what playing games with children does to them, which aligns with the context. Therefore, the query is relevant to the retrieved context."\}\\
\midrule
\textbf{GPT-3.5 Example 2}\\
\midrule
\textbf{Query}: What could a solicitor do to warn of upcoming litigation?\\
\textbf{Document}: Table 3.3 PREPARED Checklist Process for Informed Communication Safety in the Office Setting an ethical right, it is a therapeutic necessity, according to Leape. Several programs are under way to test the assertion that disclosure and apology can decrease the likelihood of litigation. COPIC, a Colorado medical insurance company, found that full disclosure results in small early settlements and dramatically reduced law suits and payouts (59). or lead such a discussion in the absence of those skills. Disclosure and apology cause apprehension for physicians—particularly in the discipline of obstetrics and gynecology where litigation adversely affected practice patterns (defensive medicine) and heightened a reluctance to disclose medical errors for fear of litigation (7). It is interesting to note that open disclosure overall generated less litigation than failing to disclose, and the growth of compensation with disclosure seems to add to this decrease (8– 12). Apology raises particular anxiety about implying culpability and inciting litigation, so help with framing an apology is always appropriate. The obligation of trust (fiduciary relationship) that we have with our patients is part of the healing aspects of medicine—and we owe it to our patients and to ourselves to develop the robust curricula and support at all levels of medicine to make disclosure the step toward solution and healing that it can be for both physician a lack of empathy and understanding for the patient, whereas concern or anxiety in the surgeon’s voice is often positively related to expressing concern and empathy. General and orthopedic surgeons whose tone of voice was judged to be more dominant were more likely to have been sued than those who sounded less dominant. \ldots\\
\textbf{Output}: \{'score': 1, 'feedback': "The query is highly relevant to the context. The context discusses the importance of disclosure and apology in reducing the likelihood of litigation. It mentions that full disclosure results in small early settlements and reduced lawsuits and payouts. It also highlights the fear of litigation as a barrier to error disclosure. Therefore, a solicitor could play a crucial role in warning of upcoming litigation by advising healthcare professionals on the importance of disclosure and apology, and providing guidance on how to handle errors and communicate with patients."\}\\
\bottomrule
\end{tabular}
\end{subtable}
\end{table*}

\begin{table*}\ContinuedFloat
\centering
\footnotesize
\begin{subtable}{\textwidth}
\centering
\caption{GPT-4}\label{tab:gpt4_example}
\begin{tabular}{p{12cm}}
\toprule
\textbf{GPT-4 Example 1}\\
\midrule
\textbf{Query}: What can you use hear other people and be heard yourself?\\
\textbf{Document}: In many situations, including lectures and the theater, hearing-impaired persons benefit from assistive devices that are based on the principle of having the speaker closer to the microphone than any source of noise. Assistive devices include infrared and frequency-modulated (FM) transmission as well as an electromagnetic loop around the room for transmission to the individual’s hearing aid. Hearing aids with telecoils can also be used with properly equipped telephones in the same way. Adapted and updated from Weiss CE, Lillywhite HE: Communication Disorders: a handbook for prevention and early detection, St Louis, 1976, Mosby. the results of testing or initiation of speech-language therapy, parents should be advised to speak slowly and clearly to the child (and avoid baby talk). Parents and older siblings should read frequently to the speech-delayed child. Speech disorders include articulation, fluency, and resonance disorders. Articulation disorders include difficulties producing sounds in syllables or saying words incorrectly tothe point that other people cannot understand what is beingsaid. Fluency disorders include problems such as stuttering,the condition in which the flow of speech is interrupted byabnormal stoppages, repetitions (st-st-stuttering), or prolonging sounds and syllables (ssssstuttering). Resonance or voice disorders include problems with the pitch, volume, or quality of achild’s voice that distract listeners from what is being said. Empathize with the patient and/or family’s feelings. Explore by asking open-ended questions. Delineate for the patient and the family the next steps, including additional tests or interventions. \ldots\\
\textbf{Output}: \{'score': 1.0, 'feedback': "The context provided discusses various assistive devices and methods for helping individuals with hearing impairments. It mentions devices such as infrared and FM transmission systems, electromagnetic loops, and hearing aids with telecoils that can be used in conjunction with telephones. Additionally, it touches on cochlear implants and their benefits for speech recognition and voice modulation. The query asks about what can be used to hear others and be heard, which is directly relevant to the assistive devices and methods described in the context for hearing-impaired individuals. Therefore, the context is highly relevant to the query."\}\\
\midrule
\textbf{GPT-4 Example 2}\\
\midrule
\textbf{Query}: When a person is living on the streets they likely what?\\
\textbf{Document}: and motor activity are characteristic of the disease. Ultimately, a deteriorated and dilapidated state occurs, which in the extreme results in an unkempt and malnourished state with which the public unfortunately associates schizophrenia. Individuals of this type roam the streets and live in appalling conditions on the fringes of society where they are subject to the criminal behavior of others. V60.0(259.0) Homelessness This category should be used when lack of a regular dwelling or living quarters has an impact on an individual’s treatment or prognosis. An individual is considered to be homeless if his or her primary nighttime residence is a homeless shelter, a warming shelter, a domestic violence shelter, a public space (e.g., tunnel, transportation station, mall), a building not intended for residential use (e.g., abandoned structure, unused factory), a cardboard box or cave, or some other ad hoc housing situation. V60.1 (259.1) Inadequate Housing Intervention is needed for caregivers who disclose IPV. It is appropriate to show concern and to provide available community resources. It is important to assess for the safety of the victim and the children. In some states, physicians are mandated to report IPV. Information for families that provides details about community resources and state laws is helpful. Youth violence is a leading cause of pediatric mortality in the United States. Homicide is the second leading cause of death for all children 1 to 19 years of age. Each year nearly 6000 children, primarily adolescents, are victims of homicide, and 4000 How are things at home and at school? Who lives with you? How do you get along with your family members? \ldots\\
\textbf{Output}: \{'score': 1.0, 'feedback': "The context provided discusses the conditions associated with homelessness, including the lack of a regular dwelling, living in shelters, public spaces, or inadequate housing. It also touches on the impact of homelessness on treatment or prognosis of individuals, and the risks they face, such as being subject to criminal behavior. The query asks about the likely conditions of a person living on the streets, which is directly relevant to the context describing the state of homelessness and the associated risks and living conditions. Therefore, the query is highly relevant to the provided context."\}\\
\bottomrule
\end{tabular}
\end{subtable}
\end{table*}

\begin{table*}
\centering
\caption{In-context examples used for generating synthetic queries.}
\label{tab:in_context}
\begin{tabular}{p{12cm}}
\toprule
\textbf{Textbooks}\\
\midrule
\{
"question": "A 45-year-old woman with metastatic breast cancer presents with acute-onset dyspnea and chest pain. She has been receiving paclitaxel chemotherapy for the past 3 months. Chest X-ray reveals pleural effusion. Which of the following mechanisms best explains the mode of action of paclitaxel?",
"options": \{
"A": "Inhibition of proteasome",
"B": "Hyperstabilization of microtubules",
"C": "Generation of free radicals",
"D": "Cross-linking of DNA"
\},
"answer": "B",
\}

\{
"question": "A 25-year-old woman presents to her gynecologist for birth control counseling. She has no significant past medical history. She expresses interest in using an intrauterine device (IUD) as her preferred method. Her vital signs are: blood pressure 120/80 mm Hg, pulse 70/min, and respiratory rate 16/min. She is afebrile. Physical examination is unremarkable. Which of the following conditions would be a contraindication to the placement of a levonorgestrel-releasing IUD in this patient?",
"options": \{
"A": "A history of severe migraines with aura",
"B": "Known uterine fibroids",
"C": "History of endometrial cancer",
"D": "Active or recent history of sexually transmitted infection (STI)"
\},
"answer": "C",
\}\\
\midrule
\textbf{PubMed} \\
\midrule
\{
"question": "Is the use of magnetic resonance imaging (MRI) superior to computed tomography (CT) in diagnosing soft tissue injuries?",
"options": \{
"A": "yes",
"B": "no",
"C": "maybe",
\},
"answer": "B",
\}\\

\{
"question": "Does the administration of statins correlate with a reduced risk of cardiovascular events in diabetic patients?",
"options": \{
"A": "yes",
"B": "no",
"C": "maybe",
\},
"answer": "A",
\}\\

\{
"question": "Can telemedicine effectively replace in-person consultations for routine follow-up appointments in managing chronic diseases?",
"options": \{
"A": "yes",
"B": "no",
"C": "maybe",
\},
"answer": "C",
\}\\
\bottomrule
\end{tabular}
\end{table*}

\begin{table*}
\centering
\caption{Examples of synthetic queries.}
\label{tab:synthetic_query}
\begin{tabular}{m{12cm}}
\toprule
\textbf{Synthetic Queries} generated from \textit{Textbooks} Corpus\\
\midrule
(1) A study conducted on hospital resident physicians revealed that working for more than 24 consecutive hours increases the risk of which of the following? \\
(2) During which phase of the menstrual cycle does the corpus luteum produce high levels of progesterone, estradiol, and inhibin? \\
(3) A 30-year-old female presents with severe headache, visual disturbances, and signs of intracranial or orbital extension. Computed tomography scanning confirms the diagnosis of a brain abscess. Which of the following is the most appropriate management approach for this patient? \\
(4) A patient presents with septic shock and hypoperfusion. Which of the following cytokines is one of the earliest to be released in response to injurious stimuli? \\
(5) A 35-year-old man with a history of multiple sclerosis presents with new neurological symptoms. A CT scan of the brain reveals contrast-enhanced ring lesions that simulate abscess or tumor. Which of the following imaging modalities is preferred for better sensitivity in detecting cerebral lesions in patients with multiple sclerosis? \\
\midrule
\textbf{Synthetic Queries} generated from \textit{PubMed} Corpus\\
\midrule
(1) Is eugenol known for its anti-inflammatory properties?\\
(2) Do DAT cells in the rat junctional epithelium possess stress fibers composed of actin filaments?\\
(3) Is splenohepatoplasty a viable method for hepatic revascularization in rats?\\
(4) Is there an independent relationship between systemic inflammation and fragmented QRS complexes in patients with stable angina pectoris?\\
(5) What type of conformation do the furanose rings exhibit in the crystal structure of the compound described?\\
\bottomrule
\end{tabular}
\end{table*}

\end{document}